%% file: main.tex
\RequirePackage{fix-cm}
\documentclass[smallextended, final]{svjour3}       
\journalname{Data Mining and Knowledge Discovery}
\smartqed  

\usepackage[square,sort,comma,numbers]{natbib}
\usepackage{comment}
\usepackage{hyperref}
\usepackage{booktabs}

\usepackage{graphicx}
\usepackage{color,soul}
\usepackage{subcaption}
\usepackage{url}            
\usepackage{amsfonts}       
\usepackage{nicefrac}       
\usepackage{tabularx,booktabs}
\usepackage[linesnumbered,ruled,vlined]{algorithm2e}
\usepackage{enumitem}
\everymath{\displaystyle}

\newcommand{\ts}{\textsuperscript}
\usepackage{xcolor}

\begin{document}

\title{
Robust Explainer Recommendation for Time Series Classification
}


\titlerunning{Robust Explainer Recommendation for TSC}

\author{Thu Trang Nguyen \and Thach Le Nguyen \and Georgiana Ifrim}

\institute{Thu Trang Nguyen, Thach Le Nguyen, and Georgiana Ifrim are with the School of Computer Science, University College Dublin, Ireland. \\
\email{thu.nguyen@ucdconnect.ie; \{thach.lenguyen, georgiana.ifrim\}@ucd.ie}}

\date{Received: date / Accepted: date}
\maketitle

\input{0abstract}

\input{1intro.tex}
\input{2relwork.tex}

\input{3background-definition}

\input{3method.tex}

\input{4experiment.tex}

\input{5discussion.tex}
\input{5.1Recommendations}
\input{6conclusion.tex}

\begin{acknowledgements}
We would like to thank the anonymous reviewers for their detailed and constructive feedback. We would also
like to gratefully acknowledge the work by researchers at University of California Riverside, USA (especially
Eamonn Keogh and his team) for their effort in collecting, updating and making available the UCR time series classification
benchmarks. We want to thank all researchers in time series classification and explainable AI who have made their data, code
and results open source and have helped the reproducibility of research methods in this area. This work was funded by Science Foundation Ireland through the SFI Centre for Research Training in Machine Learning (18/CRT/6183), the Insight Centre for Data Analytics (12/RC/2289\_P2) and the VistaMilk SFI Research Centre (SFI/16/RC/3835).  
\end{acknowledgements}

\bibliography{references}

\newpage
\input{0appendix}

\end{document}

%% file: 0abstract.tex
\begin{abstract}
Time series classification is a task which deals with temporal sequences, a prevalent data type common in domains such as human activity recognition, sports analytics and general sensing. In this area, interest in explainability has been growing as explanation is key to understand the data and the model better. Recently, a great variety of techniques (e.g., LIME, SHAP, CAM) have been proposed and adapted for time series to provide explanation in the form of \textit{saliency maps}, where the importance of each data point in the time series is quantified with a numerical value. However, the saliency maps can and often disagree, so it is unclear which one to use. 
This paper provides a novel framework to \emph{quantitatively evaluate and rank explanation methods for time series classification}. 
We show how to robustly evaluate the informativeness of a given explanation method (i.e., relevance for the classification task), and how to compare explanations side-by-side. \textbf{\emph{The goal is to recommend the best explainer for a given time series classification dataset.}}
We propose AMEE, a Model-Agnostic Explanation Evaluation framework, for recommending saliency-based explanations for time series classification. In this approach, data perturbation is added to the input time series guided by each explanation. 
Our results show that perturbing discriminative parts of the time series leads to significant changes in classification accuracy, which can be used to evaluate each explanation. To be robust to different types of perturbations and different types of classifiers, we aggregate the accuracy loss across perturbations and classifiers.
This novel approach allows us to recommend the best explainer among a set of different explainers, including random and oracle explainers. We provide a quantitative and qualitative analysis for synthetic datasets, a variety of time-series datasets, as well as a real-world case study with known expert ground truth.

\keywords{Time Series Classification \and Explainable AI \and Explanation Recommendation 
 \and Trustworthy AI}
\end{abstract}

%% file: 1intro.tex
\section{Introduction}

The last decade witnessed a rapid integration and increased impact of machine learning in  everyday life.
Machine learning algorithms work well in many applications and grow ever more complex with models having millions of parameters \cite{Devlin2018Bert:Understanding, Brown2020LanguageLearners}. Data quality is key and checking against data leakage or bias is important to enable robust models. 
However, we are still behind in explaining why these algorithms work so well and occasionally fail to perform well, and what is in the data that leads multiple classifiers to predict a certain class \cite{Goodfellow2015ExplainingExamples}. The evaluation of explanation methods is still an open problem. 
\emph{While we have many new explanation methods and methodologies, it is still difficult to decide which is the best explainer for a given problem and dataset.}

This unmatched growth of complexity and explanation of many machine learning algorithms and data, including those for time series, undermines application of these technologies in critical, human-related areas such as healthcare, sports and finance \cite{Caruana2015IntelligibleReadmission,Lipton2018TheSlippery.}. As time series data is prevalent in these applications \cite{Petitjean2014DynamicClassification,Ramgopal2014SeizureEpilepsy,Avci2010ActivitySurvey}, Time Series Classification (TSC) algorithms often call for reliable explanations \cite{Bostrom2018TheIntelligence,IsmailFawaz2019AccurateNetworks}. This explanation is usually presented in the form of \textit{feature importance} or as \textit{saliency weights} \cite{Adebayo2018SanityMaps}, highlighting the parts of the time series which are informative  for the classification decision. Saliency-based explanations were shown to be useful to find important motifs in data \cite{LeNguyen2019InterpretableRepresentations, IsmailFawaz2019AccurateNetworks}, and as a starting point to prioritise features for further investigation in counterfactual explanation methods \cite{Delaney2021Instance-basedClassification}.

\begin{figure}[hbt!]
\vspace*{5pt}
  \centering
  \includegraphics[width=1\textwidth]{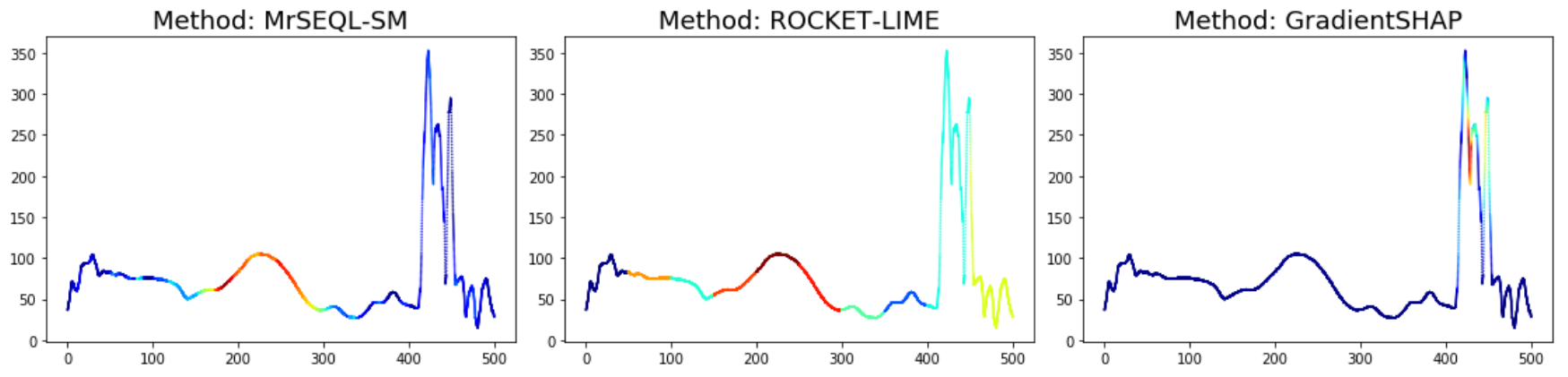}
  \caption{Saliency map explanation is a vector of feature importance weights overlaid over the original time series, where each point in the time series is coloured according to its importance.  The saliency is obtained by classifying a motion time series using different classifiers and explainers. The most discriminative parts according to the explanation method are colored in deep red, and the non-discriminative parts are colored in deep blue. }
  \label{fig:intro}
\end{figure}

Recent efforts in designing intrinsically explainable machine learning algorithms, as well as building post-hoc saliency-based explainers for black-box algorithms, have gained significant attention \cite{Zhou2016LearningLocalization,Selvaraju2017Grad-cam:Localization,Ribeiro2016WhyClassifier,Lundberg2017APredictions}.
Most works focus on explaining one particular classification algorithm, with a lot of emphasis on deep learning methods \cite{boniol2022dcam,Zhou2016LearningLocalization, Selvaraju2017Grad-cam:Localization, Sundararajan2017AxiomaticNetworks,Smilkov2017Smoothgrad:Noise, springenberg2015striving}. Often we are faced with a set of saliency maps for explanation, coming either from domain experts or from a diverse set of classifiers. In particular, for time series classification, a variety of classifiers are required for high accuracy, depending on the application domain \cite{Bagnall2016Time-seriesEnsembles, middlehurst2023bake}. Each of these classifiers may be tied in accuracy, can be explained with different methods (e.g., LIME or SHAP), and often the resulting explanations disagree, e.g., pointing to different parts of a time series as most relevant for a predicted class. 
We thus face the challenge: \textit{How to assess and objectively compare many explanation methods?}  
In other words, if two or more explanation techniques give different explanations (i.e., two different saliency maps coming from the same classifier or different classifiers, Figure \ref{fig:intro}), \textit{which explanation is best for our task?} In this paper, we focus on the time series classification task and propose a methodology appropriate for time series. Some of the ideas we investigate are relevant to recommending explainers beyond the TSC task, but this is beyond the scope of this paper.

We propose a methodology to compute a standardized evaluation measure, which enables  quantitative comparison and ranking of explainers (Table \ref{tab:desired-outcome}). From the application users' perspective, having this recommendation can 
support short-listing of useful explanations for further analysis and optimisation \cite{Doshi-Velez2017TowardsLearning}. 
At the very least, we want to know that a given explanation is better (more informative) than a random explanation, and in general we want to be able to select the best explainer for a given dataset.


\begin{table}[!hbt]
    \vspace{5pt}
    \centering
    \begin{tabular}{lcl}
    \toprule
            & \begin{tabular}[c]{@{}c@{}}Explanation Power\\ {[}0-1{]}\end{tabular} & \begin{tabular}[c]{@{}c@{}}Method Ranking\\ {[}1-5{]}\end{tabular} \\
            \midrule
Oracle      &   1.00    & \hspace{0.7cm} 1(best)     \\
MrSEQL-SM   &    0.90   &   \hspace{0.7cm}   2        \\
ROCKET-LIME & 0.56      & \hspace{0.7cm} 3             \\
ResNet-GradientSHAP    & 0.06     & \hspace{0.7cm} 4               \\
Random      &    0.00   &   \hspace{0.7cm} 5 (worst)                                                     \\
\bottomrule
\end{tabular}
\smallskip
    \caption{Outcome of AMEE: a measure to evaluate multiple explanation methods. \textbf{Explanation Power} measures the informativeness of each explanation, taking values from 0 to 1, where 0 is worst and 1 is best.} 
\label{tab:desired-outcome}
\end{table}
 
In this paper, we present \textit{\textbf{A M}odel-agnostic framework for \textbf{E}xplanation \textbf{E}valuation for Time Series Classification} (\textbf{AMEE}).
Specifically, we focus on explanations in the form of a \textit{saliency map} and consider their  informativeness within a defined computational scope, in which a more informative explanation means a higher capacity to influence classifiers to  identify a class. We show that the saliency-guided perturbation of discriminative subsequences results in a reduced accuracy of classifiers. The higher the impact of a perturbation, the more informative are the perturbed time series subsequences. Estimation of this impact, measured by a committee of highly accurate \textbf{referee classifiers}, can reveal the informativeness of the explanation. This is the key idea behind AMEE, a post-hoc approach which uses a set of classifiers and explainers to recommend the best explainer for a given time series classification dataset.



Our work addresses an overlooked area of research: \emph{robust comparison and ranking of multiple explanation methods for time series classification}. Our main contributions are:
\begin{itemize}
    \item A robust, model-agnostic, 
    ensemble-based explanation evaluation framework.
    First, we leverage the use of multiple data perturbation strategies to create explanation-guided noisy data. Using synthetic data, we empirically show that applying multiple data perturbation strategies is particularly useful when the data is hard-to-classify, as such data is often more sensitive to the data perturbation type. We also show that a committee of referee classifiers is useful to reduce the potential bias that one single referee classifier may  have. Our experiments demonstrate that a committee approach involving multiple types of data perturbations and multiple classifiers leads to explanation evaluation and ranking that better agrees with the explanation ground truth (synthetic data) and domain expert ground truth (real data).
    \item A standardised evaluation measure (\textbf{Explanation Power}) that is comparable across different explanation methods, referee classifiers and datasets.
    \item An empirical study on both synthetic and real datasets with recent state-of-the-art time series classifiers and explanation methods. We verify  the evaluation methodology with annotated, real datasets. All data, code and detailed results are  available\footnote{Data and code are available at: \url{https://github.com/mlgig/amee}}.
    
\end{itemize}

In the next sections  we review related Explainable AI research including both time series specific and general methods (Section \ref{sec:relwork}). We then define related concepts (Section \ref{sec:background}) and describe our proposed solution (Section \ref{sec:method}). We discuss experiments on both synthetic and real time series datasets,  with detailed case studies (Section \ref{sec:experiment}). We discuss important considerations for practitioners when using AMEE to evaluate and recommend explanations for time series classification in Section \ref{sec:recommendation}. Finally, we summarize  our results and discuss future work in Section \ref{sec:conclusion}.

%% file: 2relwork.tex
\section{Related Work}
\label{sec:relwork}

\subsection{Explanation Methods for Time Series Classification}

As deep learning has achieved high performance in machine learning domains such as computer vision \cite{Szegedy2015GoingConvolutions, Krizhevsky2012ImagenetNetworks}, the research community started to develop techniques to explain these black-box models to understand why they work so well  \cite{Zhou2016LearningLocalization, Selvaraju2017Grad-cam:Localization, Sundararajan2017AxiomaticNetworks,Smilkov2017Smoothgrad:Noise, springenberg2015striving}. These explanations are in the form of a saliency map, visualizing the important pixels in an  image by computing a saliency weight for each pixel (a type  of feature importance). This saliency map, combined with the original image, can reveal whether a black-box model focuses on the correct area of the image and \textit{explain} the model in a visually friendly way. 

Explainable AI (XAI) methods for Time Series Classification have advanced in parallel with general XAI progress \cite{Theissler2022}. Although recent works on \textit{instance-based} methods, such as factual and counterfactual explanations \cite{Guidotti2020,Delaney2021Instance-basedClassification,WangZhendongandSamsten2021LearningRepresentations} have become popular, the majority of explanation methods exist in the form of \textit{saliency maps} \cite{kokhlikyan2020captum,mishra-2017-local,Parvatharaju2021,Rooke2017,Guilleme2019, Zhou2016LearningLocalization, Smilkov2017Smoothgrad:Noise}, where a map visualizes the importance weight vector $w$ and highlights the discriminative areas of a time series for the classification task. These \textit{saliency-based explanations} can either be extracted directly from the classifier (intrinsic explanation), or indirectly by applying a post-hoc explanation method to the black-box classifier (post-hoc explanation).     

\subsubsection{Intrinsic Explanation}
\emph{Explanation from MrSEQL Time Series Classifier.} MrSEQL 
\cite{LeNguyen2019InterpretableRepresentations} is a time series classification algorithm that is intrinsically explainable.
The algorithm converts the numeric time series vector into strings, e.g., by using the SAX \cite{Lin2007ExperiencingSeries} transform with varying parameters to create multiple symbolic representations of the time series. The symbolic representations are then used as input for SEQL \cite{Ifrim2011BoundedSpace}, a  sequence learning algorithm, to select the most discriminative subsequences for training a classifier using logistic regression. The symbolic features combined with the classifier weights learned by logistic regression make this classification algorithm explainable. For a time series, the explanation weight of each data point is the accumulated weight of the SAX features that it maps to. These weights can be mapped back to the original time series to create a saliency map to highlight the time series parts important for the classification decision. We call the saliency map explanation obtained this way, MrSEQL-SM.
For using the weight vector from MrSEQL-SM, we take the absolute value of weights to obtain a vector of non-negative weights.
Figures \ref{fig:coffee-example} and \ref{fig:gunpoint-example} show an example of the saliency map explanation obtained directly from the MrSEQL classifier weights, for  the Coffee and GunPoint datasets from the UCR Archive \cite{Dau2018TheArchive}.

\paragraph{Explanation from a Generic, White-box Classifier.}
A generic, white-box classifier such as Logistic Regression or Ridge Regression has been the primary source of providing feature importance (by using the learned model weights), especially for tabular data \cite{HosmerJr2013AppliedRegression}. These classifiers and their explanations are computationally cheap and can be useful for time series data \cite{Frizzarin2023Classification2022}. 

\subsubsection{Posthoc Explanation}

\emph{Gradient-based Explanation.}
This approach uses the gradients from a trained deep neural network to infer explanations. 
Notable methods are Integrated Gradient \cite{Sundararajan2017AxiomaticNetworks}, GradientSHAP \cite{Lundberg2018ExplainableSurgery}, GradCAM \cite{Selvaraju2017Grad-cam:Localization}, CAM \cite{Zhou2016LearningLocalization}. For time series classification and explanation, the most common classifier is ResNet \cite{IsmailFawaz2019DeepReview} combined with some of the explanation methods mentioned.

\begin{figure}[hbt!]
\vspace*{5pt}
  \centering
  \includegraphics[width=0.8\textwidth]{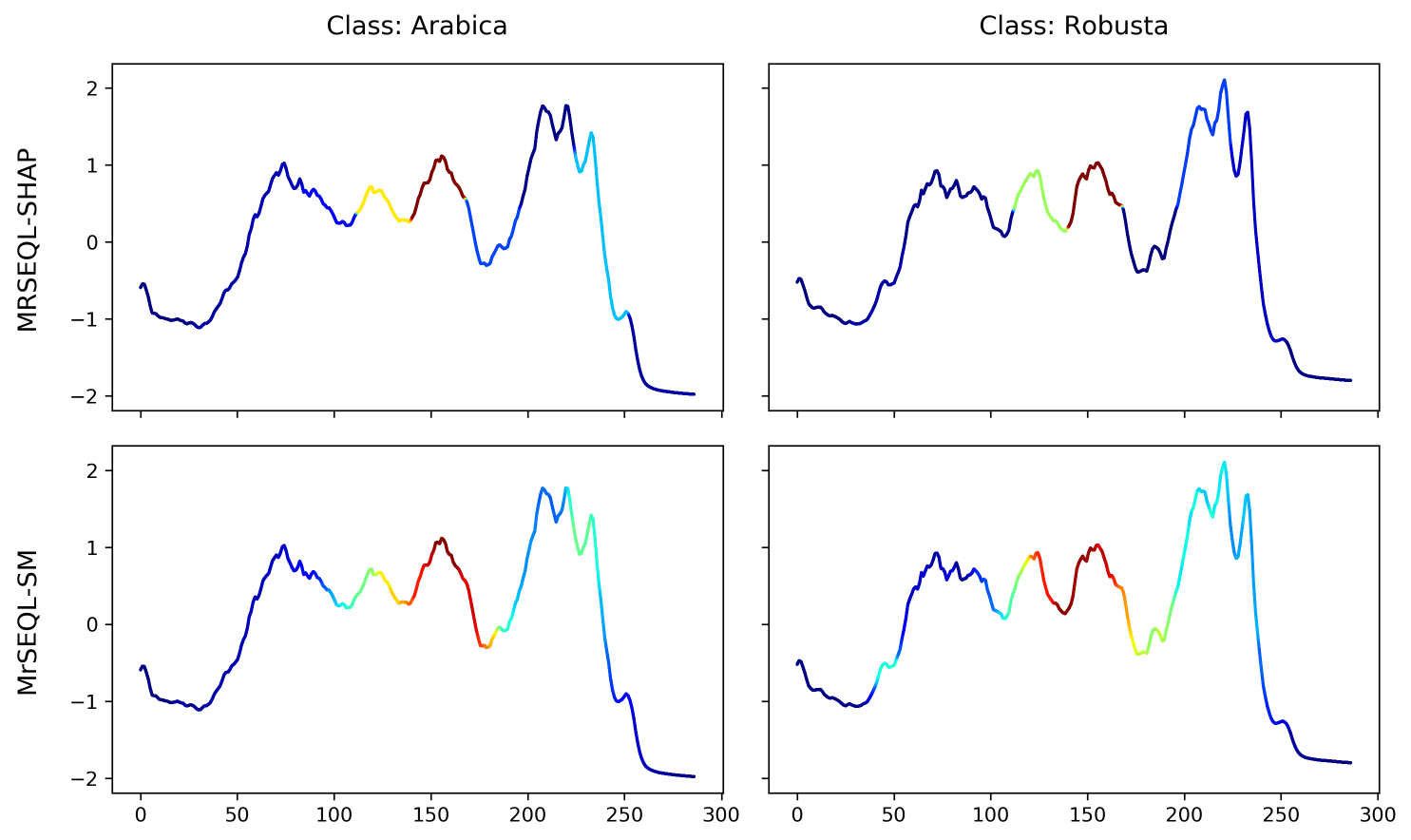}
  \caption{
  Saliency map from two explanation methods on two examples from the Coffee dataset: the bottom row is an explanation from MrSEQL Classifier (intrinsic explanation); the top row is an explanation from SHAP, a post-hoc explanation method based on MrSEQL Classifier.
  } \label{fig:coffee-example}
\end{figure}

\begin{figure}[hbt!]
\vspace*{5pt}
  \centering
  \includegraphics[width=0.8\textwidth]{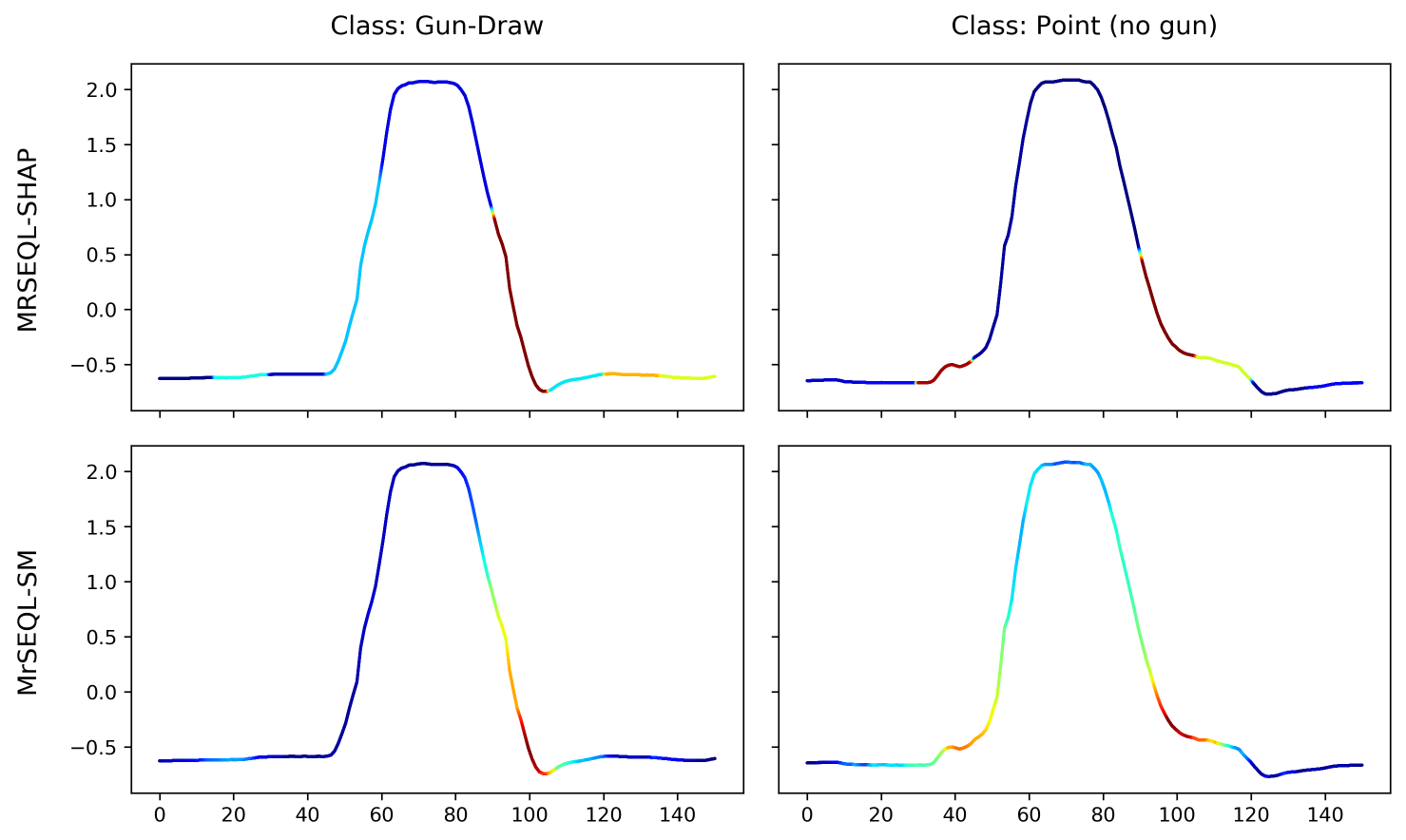}
  \caption{Saliency map from two explanation methods on two examples of GunPoint dataset: the top bottom row is explanation from MrSEQL Classifier (intrinsic explanation); the top row is explanation from SHAP, a post-hoc explanation method based on MrSEQL Classifier.}
  \label{fig:gunpoint-example}
\end{figure}

\paragraph{Perturbation-based Explanation.}
This type of methods infuse noise into the data to create data variations and to infer the degree of data point  importance \cite{CASTRO20091726, ABANDA2022109366}.
Notable methods are Feature Occlusion \cite{Suresh2017ClinicalNetworks} and LIME \cite{Ribeiro2016WhyClassifier}.
One of the most popular post-hoc explanation methods is SHAP \cite{Lundberg2017APredictions}  - a unique way to explain any machine learning model using a game theoretic approach, in which all feature coalitions are evaluated. 
Feature importance is then calculated using the classic Shapley value \cite{Strumbelj2014ExplainingContributions}.
Figures  \ref{fig:coffee-example} and \ref{fig:gunpoint-example} show  an example of the saliency map explanation obtained by applying SHAP to the MrSEQL classifier to get a post-hoc explanation for each of the Coffee and GunPoint datasets from the UCR Archive \cite{Dau2018TheArchive}.

\subsection{Quantitative Evaluation of Saliency-based Explanation}

Quantitative evaluation of explanations for time series data was a relatively untouched topic until recently. Unlike image and text, time series data often do not have annotated ground truth explanation; hence, it remains a challenge to determine whether a saliency-based explanation is correct. Approaches to benchmark and evaluate faithfulness of recent explanation methods overcome this problem by using synthetic datasets with assigned ground-truth \cite{Ismail2020BenchmarkingPredictions, Crabbe2021ExplainingMasks}. Other research ventures into real datasets, yet these  efforts focus on examining explanations by a single classifier  \cite{Crabbe2021ExplainingMasks} or averaging a non-comparable metric across multiple datasets \cite{Schlegel2019TowardsSeries}. The approach in  \cite{Guidotti2021EvaluatingTruth} uses  a white-box classifier to get a pseudo ground-truth explanation (a) and  evaluates a post-hoc, localized explanation method (b) by  estimating cosine distance between (a) and (b). However, this method assumes that the white-box classifiers can always produce explanations of ground-truth quality. We show in our experiments that this is not the case.
Notably, \cite{Schlegel2019TowardsSeries,Nguyen2020AClassification,Agarwal2021RankingClassification}  
propose methods to quantify explanation methods, however, there are a few problems with the comparison: the use of a single perturbation type is problematic as it cannot always distinguish between explanations, the metric used (change in accuracy) is not comparable across the selected datasets, the individual effect is not separated (only average change in accuracy is reported), and there is no discussion involving explanation ground-truth.
Additionally, there is little discussion in previous work about the impact of the classifier(s) accuracy on evaluating the explanation methods that are based on those  classifier(s). This is an important point, as the evaluation can only be trusted if the classifier(s) are reliable. Furthermore, there are cases where multiple classifiers have high accuracy and are tied in this regard, but the explanations obtained from the classifiers may disagree, and in some cases could be ranked worse than a random explanation. Hence it is not clear which classifier and explanation to select  in such cases.

%% file: 3background-definition.tex

\section{Background and Definitions}
\label{sec:background}

\subsection{Time Series \& Time Series Dataset}
A time series $X$ = [$x_0$,$x_1$,...,$x_{l-1}$], $x_i \in \mathbb{R}$, is a sequence of $l \in \mathbb{N}$ real values that are recorded values of a synthetic or real process. In this definition, $l$ is also called the number of time steps or the length of the time series $X$, and $x_i$ are the data points or time points.

A time series dataset $D$ consists of $n \in \mathbb{N}$ time series of equal length $l$ that are recorded from a single process. If the time series are not of equal length, it is common to pad with zeroes or use resampling to bring them to equal length.

\subsection{Saliency-based Explanations for Time Series}

In the context of this paper, we only consider explanations in the form of saliency maps. A saliency map to explain time series $X$ is a vector of numerical weights $M = [w_0, \dots ,w_{l-1}]$ where $w_i \in \mathbb{R}$ and $l$ is the length $X$. The value $w_i$ implies the importance (or saliency) of the time point $i$ in the process of prediction making for $X$. This vector can be obtained from annotation (by a human) or computed by an explanation method. The explanation method can come from a white-box classification model (intrinsic explanation) or a black-box classifier coupled with a post-hoc explanation method (post-hoc explanation). The weights $w_i$ are typically rescaled to $[0,1]$.

\subsubsection{Random Explanation}
For sanity checks, we use saliency maps generated through random sampling as a lower bound on explanation quality. Here, the weights $w_i$ are drawn from a random uniform distribution. Like a dummy classifier, this random explanation serves as a baseline for any reasonable explanation method, i.e., they all should be better than random guessing.
Nonetheless, there are situations where a random explanation outperforms a method-based explanation. Specifically, when a method-based explanation highlights non-discriminative parts, or fails to identify any discriminative parts, that explanation can be considered worse than a random explanation.

\subsubsection{Oracle Explanation} In cases where explanation ground truth is available (e.g., for synthetic datasets or from domain experts), this should be the gold standard for any explanation method. We generally expect any explanation method to rank between the random and the oracle explanations.

%% file: 3method.tex
\section{Methodology}
\label{sec:method}

In this section, we describe our proposed methodology using  concepts described in Section \ref{sec:background}. Specifically, we present the blueprint of AMEE in Figure \ref{fig:process}.
The framework involves a labelled time series dataset (split into training and test datasets), a set of explanation methods to be compared, and a set of evaluating classifiers (referee classifiers). The output of the framework is the \textit{explanation power} of each explanation method (see Table \ref{tab:desired-outcome}).

\subsection{Explanation-Guided Data Perturbation}

A good saliency-based explanation for a time series should highlight its discriminative part(s) that contain class-specific information to distinguish from other classes. Data perturbation is the process of adding noise to the data by replacing selected time points in the time series. 
Explanation-guided data perturbation uses a saliency-based explanation to determine the specific time points of the time series to be perturbed.
As a result, the more informative the explanation, the higher the decrease in classifier accuracy is expected, because that perturbation removes important class-specific information in the respective time series.
Given a threshold $k$ ($0 \leq k \leq 100$), the discriminative parts of a time series of $l$ steps are segmented using the top $k$-percentiles in $M$. This is a set of $k * l / 100$ time steps that have the highest weights in the saliency map $M$. 
Varying $k$ allows us to control the scope of the perturbation. At $k=0$, the time series is the original; at $k=10$, only 10 percent of the time steps (that are most discriminative according to the explanation) are perturbed; at $k=100$ the entire time series is perturbed.

\subsection{Referee Classifiers} 

In our work we employ a set of independent and accurate classifiers that are trained with the original training set and are used to evaluate the target explanations on the test set. This committee is formed of member classifiers that we call Referee Classifiers.
In order to evaluate the explanation methods, our framework measures the impact of each explanation-guided data perturbation on the accuracy of the referee classifiers $R$. We select the referees based on recent empirical benchmarks on TSC \cite{middlehurst2023bake}.

\subsection{Data Perturbation Strategy: Multiple Perturbations}
\label{subsec:data-perturbation-strategy}

\begin{figure}[bt!]
  \centering
  \includegraphics[width=1\textwidth]{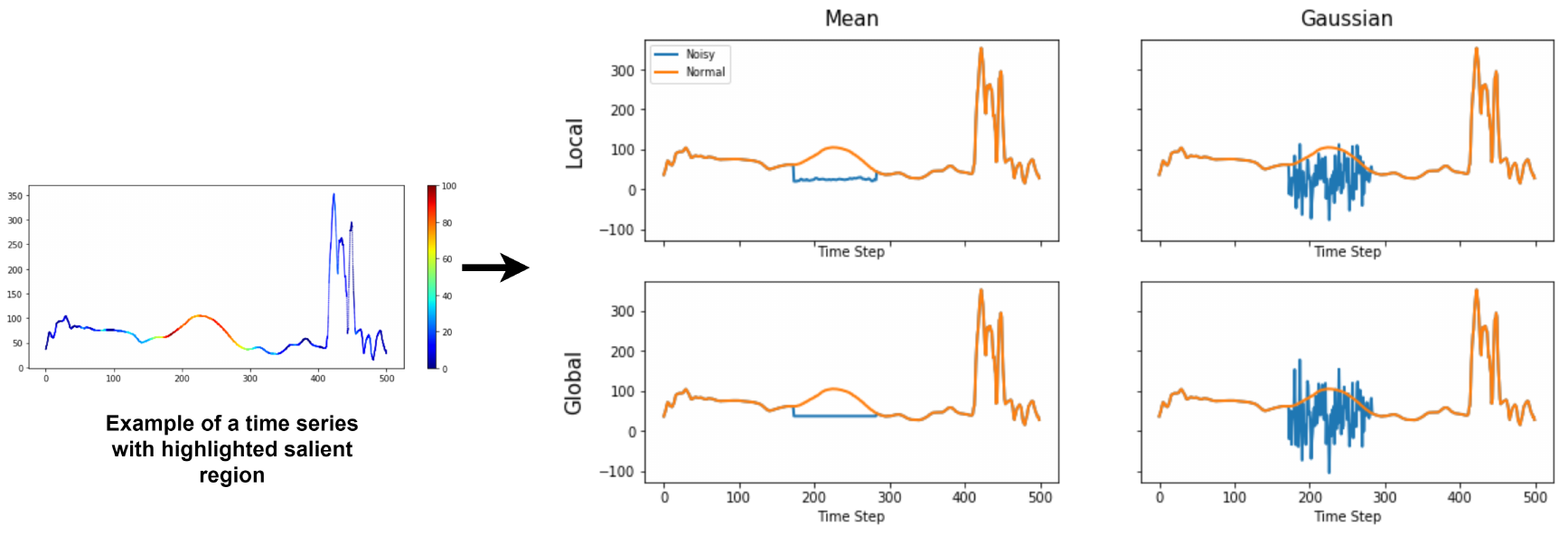}
  \caption{Time Series Data Perturbation strategy: An example time series with a known saliency map (left) is perturbed using mean or Gaussian noise using local time steps (local) or global time steps (global) across the entire dataset on its most discriminative region (in this example we perturb the top 20\% values according to the highest saliency weights).  
  }
  \label{fig:noise}
  \vspace{5pt}
\end{figure}

In Figure \ref{fig:noise} we explain and visualize four strategies to perturb the discriminative areas of a time series, as guided by a given explanation \cite{Mujkanovic2020TimeXplainClassifiers}. These strategies are either time-step dependent (local perturbation, using only the \textit{t-th} step information) or time-step independent (global perturbation), using either Gaussian-based or single value replacement. With these strategies, discriminative time steps are replaced with noisy values, either by replacing the original time series values with a patch of constant values (like a grey mask in an image) or a patch of random Gaussian noise values (like a noise mask in an image). 
Let $n$ be the number of time series in a dataset $D$, each with $l$ time steps. We want to perturb one test time series of size $1\times l$, so its \textit{t-th} value $x_t$ is replaced with a new value $r_t$. We define the global and local profile for this time step perturbation as follows.


Local perturbation:
\begin{equation}
\mu_t = \frac{1}{|n|}\sum_{x \in D}x_t;
\sigma_t^{2} = \frac{1}{|n|-1} \sum_{x \in D} (x_t - \mu_t)^2
\end{equation}

\smallskip

Global perturbation:
\begin{equation}
\mu = \frac{1}{l|n|} \sum_{i=1}^{l} \sum_{x \in D}x_i; \sigma^{2} = \frac{1}{l|n|-1} \sum_{i=1}^{l} \sum_{x \in D} (x_i - \mu)^2
\end{equation}

\smallskip

With these local-based and global-based profiles, we can define the perturbation $r_i$ accordingly. We use four perturbation strategies, two local and two global perturbation types. Local mean: $r_t^{(1)} = \mu_t$; Local Gaussian:  $r_t^{(2)} \sim \mathcal{N}(\mu_t,\,\sigma_t^{2})$;
Global mean: $r_t^{(3)} = \mu$;
Global Gaussian: $r_t^{(4)} \sim \mathcal{N}(\mu,\,\sigma^{2})$. Figure \ref{fig:noise} illustrates an example of how the four strategies effectively modify the original time series in the regions identified by the explanation weights. 
We show in our experiments that it is important to use a set of perturbation strategies, rather than a single fixed perturbation.

\begin{figure}[ht]
  \centering  
  \includegraphics[width=1\textwidth]{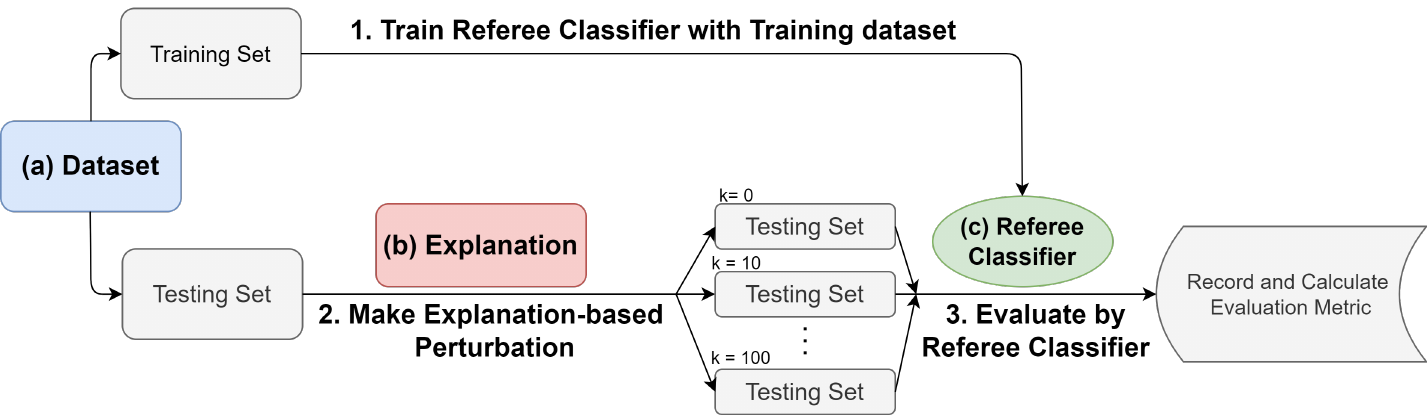}
  \caption{The AMEE evaluation framework requires 3 elements: \textbf{(a) a dataset} that requires explanation evaluation, \textbf{(b)} a set of \textbf{saliency-based explanations}, and  \textbf{(c)} a set of \textbf{referee classifiers} trained on a subset of (a). }
  \label{fig:process}
\end{figure}

\subsection{The AMEE Framework for Evaluating Explanations}

Figure \ref{fig:process} summarizes the components and steps in the AMEE framework. Our framework requires a labeled time series dataset ($D$), a set of explanations ($M$) to evaluate, and a set of referee classifiers ($R$) to be trained on a subset of the dataset. With these elements, the following steps are done to record the necessary information to calculate evaluation metrics:
\begin{enumerate}
\setcounter{enumi}{-1}
    \item Split the labeled dataset $D$ into training ($D_{train}$) and test ($D_{test}$);
    \item Train Referee Classifier(s) ($R$) with ($D_{train}$); 
    \item Use each explanation in $M$ to create a step-wise, explanation-based perturbation on $D_{test}$;
    \item Measure the accuracy of each trained referee in $R$ on these perturbed datasets $D'_{test}$. 
\end{enumerate}

The output of this process is the accuracy on the perturbed dataset $D'_{test}$ at various thresholds ($k$), serving as an indicator of how much an explanation-based perturbation impacts the referees. Significant drop in accuracy in the first few steps of the explanation-guided perturbation (e.g., at $k=10$ or $k=20$) signals that meaningful, salient data points are disturbed based on the explanation. Hence, explanations that correctly identify such salient regions are likely to be informative.

\subsection{Explanation AUC}
We measure the impact of each explanation by estimating the Area Under the Curve (AUC) of its explanation-guided perturbation. Specifically, the accuracy scores at each threshold ($k$) are translated into an Explanation-AUC ($EAUC$) using the trapezoidal rule. 

\begin{equation}
EAUC = \frac{1}{2} \mathit{\Delta}k_0 \sum_{i=1}^q (acc_{i-1} + acc_{i} )
\end{equation}

Here $\mathit{\Delta}k_0$ denotes the difference in value of each step normalized to 0-1 range ($\mathit{\Delta}k_0 = \frac{1}{100} \mathit{\Delta}k$); $q$ denotes the number of steps ($q = \frac{100}{k}$);
$acc_i$ is the accuracy at step $i$. If we perturb the dataset with $q$ steps, we will have a total of $q+1$ data points for accuracy scores. For example, if the perturbation is done in $q=10$ steps, each step will correspond to a difference of k = 10 percentage points in perturbation threshold (i.e. 0\%, 10\%, ..., 100\%). The step for $k=0$ corresponds to the original test dataset, while the step for $k=100$ corresponds to adding noise to the entire time series. 

\begin{figure}[!htb]
\centering
  \includegraphics[scale=0.2]{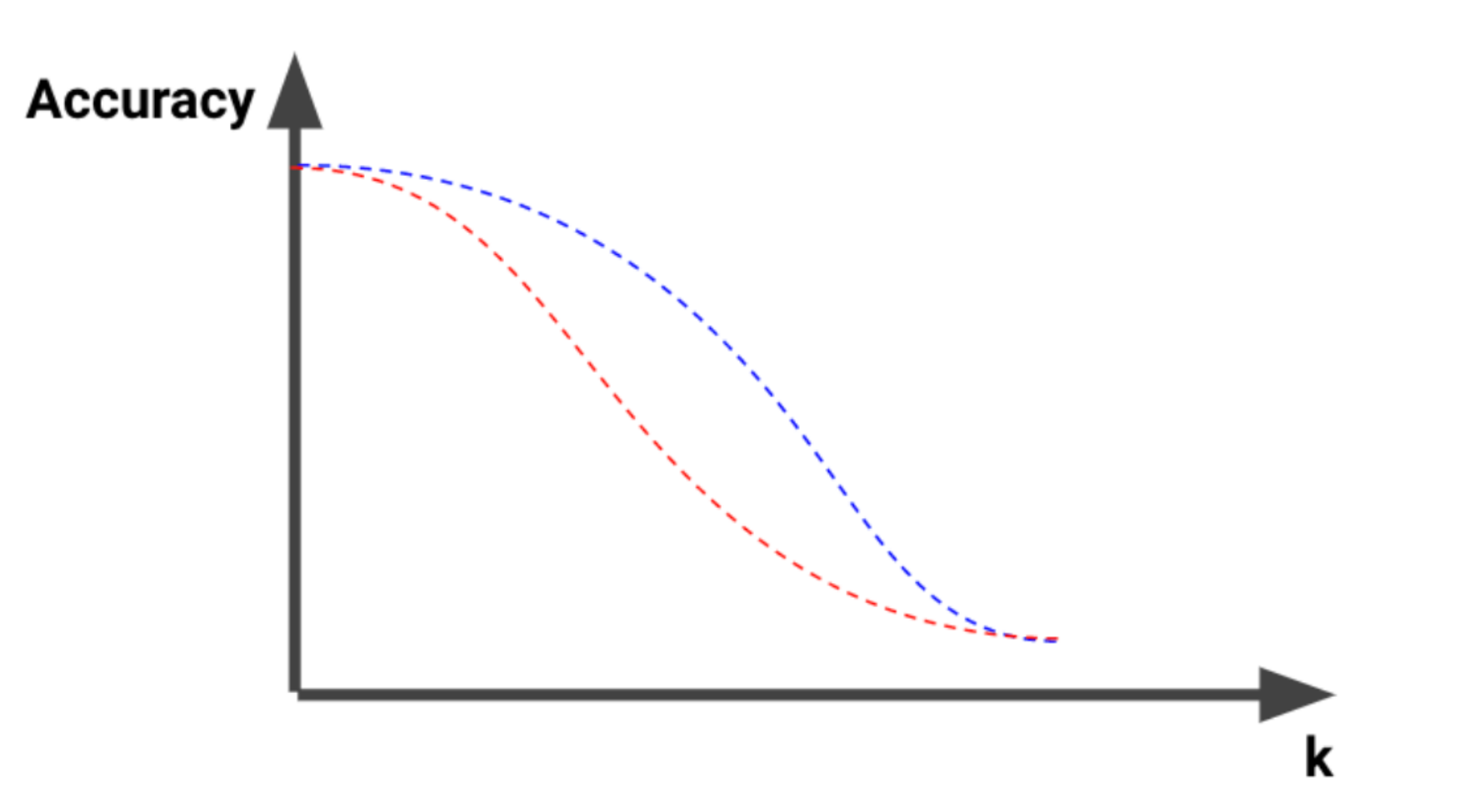}
  \caption{Changes of accuracy measured by a referee classifier among two explanation methods (red and blue) at each threshold level $k$.
  When a signal is perturbed based on a more informative explanation, this signal becomes  harder for the referee to classify correctly, leading to a more severe drop in accuracy. This impact is measured by the \textbf{Explanation AUC}, or the the area under the curve (AUC) of these changes in accuracy at different threshold $k$. The curve with lower explanation AUC (red curve) results from perturbation guided by a more informative explanation method.}
  \label{fig:explanation-auc}
\end{figure}

With this estimation, a smaller $EAUC$ means higher impact (accuracy loss) of the explanation method (Figure \ref{fig:explanation-auc}). The Explanation AUC is computed for each combination of Perturbation - Referee - Explanation (Figure \ref{fig:metric-standard}: Step 1).

\subsection{Robustness of the AMEE Framework}

Two key aspects of AMEE are aimed to make the framework more robust by employing multiple Data Perturbation strategies and multiple Referee Classifiers. Specifically, for each explanation in $M$, we use different data perturbation strategies (as described in Section \ref{subsec:data-perturbation-strategy}) to create explanation-based perturbations on the dataset $D_{test}$. Additionally, multiple referee classifiers are trained on $D_{train}$ and their accuracy is measured on the perturbed $D'_{test}$.

The various data perturbation strategies  represent different ways that salient parts of the data can be replaced with noise. Unlike image data that is standardized in RGB, time series data is more dynamic: it can belong to many different domains, collected from various sources, or preprocessed in different ways. This characteristics of time series data make it harder to use one single method to mask out a specific part of the data. Using a variety of data perturbations ensures that data is perturbed in ways that completely mask out the relevant parts of the signal. We further investigate using multiple Data Perturbation strategies in Section \ref{subsec:impact-data-perturbation-strategy}.   

Referee classifiers are used to evaluate the impact of data perturbation on the pre-trained model with the original, non-perturbed data. Thus, the evaluation by referee classifiers is dependent on the properties of these classifiers such as in-classifier data normalization, feature extraction, and feature processing. Having multiple referee classifiers can reduce potential biases introduced by using one single classifier. We analyze this characteristic and show the benefits of using multiple referee classifiers in Section \ref{subsec:impact-referees}.

\subsubsection{Standardization and Explanation Power}
\label{subsec:metric}
\begin{figure}[ht]
\centering
\includegraphics[width=1\textwidth]{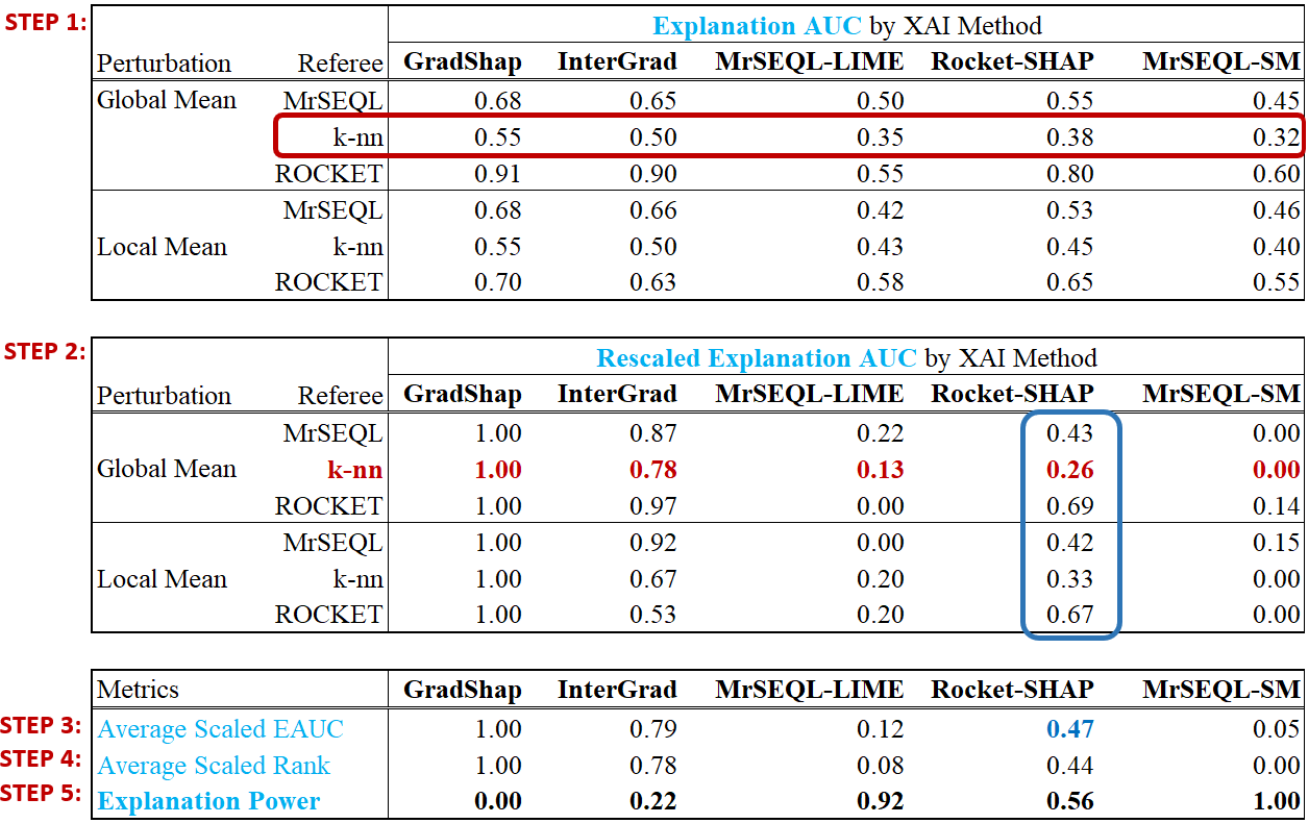}
  \caption{Measure Standardisation and Explanation Power Calculation. Example of how Explanation Power is derived in a typical evaluation assessment, involving 2 perturbation strategies (local, global) and 3 referees (MrSEQL, k-nn, ROCKET).}
  
  \label{fig:metric-standard}
\end{figure}

AMEE employs multiple perturbation strategies and multiple referee classifiers. As the EAUC measures depend on the choice of referees and perturbation strategies, they are not directly comparable. The next steps (Figure \ref{fig:metric-standard}: Step 2-5) standardize and aggregate the EAUC to compute the final output of the framework, the \textbf{Explanation Power}. 

Step 2 rescales the Explanation AUC to the same range $[0,1]$ for each row (i.e. each pair of Referee and Perturbation). 
Since each referee responds to changes in the perturbed dataset differently, this normalization is performed for each pair of Referee and Perturbation to ensure that the Explanation AUC is comparable across the different explanation methods in the evaluation.
The red highlighted row is an example. After rescaling the Explanation AUC, the \textbf{Average Scaled EAUC} is computed in Step 3. It is basically the average of each column in Step 2. 
This simple average calculation can be performed because individual Explanation AUC are already normalized to the [0,1] range and comparable across each Referee-Perturbation pair.
For example the Average Scaled EAUC of Rocket-SHAP is $(0.43 + 0.26 + 0.69 + 0.42 + 0.33 + 0.67)/6 = 0.47$. 

In Step 4, the Average Scaled EAUC is again rescaled to the range between 0 and 1. The result is the \textbf{Average Scaled Rank} (lower is better). The \textbf{Explanation Power} is simply the inverse of Average Scaled Rank ($1-$ Average Scaled Rank), i.e., higher is better.
Details of this calculation are summarised in Algorithm \ref{algo:process_data}.

\begin{algorithm}
\small

\KwIn{
Set of XAI methods $M$, set of Perturbations $T$, set of Referees $R$, set of thresholds for important area $k$, test accuracy ($acc_{M,T,R,k}$)}
\KwOut{Average Scaled Explanation AUC ($asEAUC_M$), Average Scaled Rank ($asRank_M$), Explanation Power ($ePower_M$)}
Calculate Explanation AUC ($EAUC_{M,T,R}$) using $acc_{M,T,R,k}$ \\
Calculate rescaled AUC ($sEAUC_{M,T,R}$) by Min/Max Rescaling $EAUC_{M,T,R}$ \\
Calculate Average Scaled Explanation AUC ($asEAUC_M$) of each $M$ by averaging $sEAUC_{M,T,R}$ across $R$ and $T$\\
Calculate Average Scaled Rank ($asRank_M$) of each $M$ by Min/Max Rescaling $asEAUC_M$\\
Calculate Explanation Power ($ePower_M$) by 1 - $asEAUC_M$
\caption{AMEE: Calculate Explanation Power}
\label{algo:process_data}
\end{algorithm}
\vspace{-1cm}

\ \\

%% file: 4experiment.tex
\section{Experiments}
\label{sec:experiment}

In this section, we evaluate the performance of the AMEE framework in three groups of experiments in ascending order of difficulty. In the simplest case, we want to validate AMEE with synthetic datasets with known explanation ground-truth \cite{Ismail2020BenchmarkingPredictions}. Next, we measure the performance of the framework with a diverse set of time series classification datasets from the UCR Time Series Classification Archive  covering popular domains that require explanation 
\cite{Dau2018TheArchive}. Finally, we test our framework on a real dataset and compare the result with ground-truth explanations provided by domain experts. Our experiments are repeated 5 times and the reported results are the average of these  repetitions. 

\subsection{Referee Classifiers}
We employ 5 candidates for referee classifiers in our experiment, selected based on their accuracy, speed and diversity of approach   \cite{Schafer2023WEASELClassification}: baseline 1NN-DTW  (distance-based) \cite{Cover1967NearestClassification}, MrSEQL (dictionary-based, time domain)  \cite{LeNguyen2019InterpretableRepresentations}, ROCKET (convolution-based) \cite{Dempster2020ROCKET:Kernels}, RESNET (deep learning) \cite{He2016DeepRecognition,IsmailFawaz2019DeepReview} and WEASEL 2.0 (dictionary-based, frequency domain) \cite{Schafer2023WEASELClassification}. 
As the choice of referees is a critical component in our framework, we carefully select classifiers that perform well in accuracy on all studied datasets. For a classifier to be selected in the referee committee, it has to achieve at least the average accuracy of all candidates for referee classifiers, and this number has to be higher than the theoretical accuracy achieved by a random classifier. In case the average accuracy is over 90\%, the threshold to choose referees is set to  90\%. 
By using a high accuracy threshold in cases when average accuracy is relatively high, we want to include the referees that do have high performance but slightly below the average accuracy. For example, in a theoretical case when the accuracies of 5 candidate classifiers are 0.90, 0.95, 0.97, 0.98, 0.99; we want to include all the referees as they all have relatively high performance, despite one that is  slightly below the average accuracy. For some datasets, all the classifiers are tied or very close in accuracy. 
Details of the referee accuracy are presented in the Appendix. 

\subsection{Explanation Methods}
In our experiments, we evaluate 8 popular explanation methods with  diverse properties as described in  Table \ref{tab:explanation-method}.

We use the author's implementation for LIME  \cite{Ribeiro2016WhyClassifier} and MrSEQL \cite{LeNguyen2019InterpretableRepresentations}, the \emph{captum} \cite{kokhlikyan2020captum} library for gradient-based explainers, the \emph{time-explain} library \cite{Mujkanovic2020TimeXplainClassifiers} for SHAP, and \emph{sklearn} \cite{sklearn_api} to implement the remaining classifiers and explainers. We have considered a few other recent explainers, e.g., LIMESegment \cite{sivill2022limesegment}, but they proved too slow to be feasible to run on all our datasets. Since our goal is to rank a set of given explainers, rather than promote any particular explainer, we consider this explainer set to be sufficient to validate our methodology.

\newcolumntype{Y}{>{\centering\arraybackslash}X}

\begin{table}[!hbt]
\fontsize{9pt}{9pt}\selectfont
    \centering

    \begin{tabularx}{1\textwidth}{l * {5}{Y}}
    \toprule
    Explanation & Type & Model & Explanation & Time-Series & Library \\ 
    Method & & Specific & Scope & Specific & \\
    \midrule
    GradientSHAP & Post-hoc & Yes & Global & No & \textit{captum}\\ 
            Integrated Gradient & Post-hoc & Yes & Global & No & \textit{captum} \\
            MrSEQL-LIME & Post-hoc & No & Local & No & \textit{LIME} \\ 
            ROCKET-LIME & Post-hoc & No & Local & No & \textit{LIME}\\ 
            MrSEQL-SHAP & Post-hoc & No & Local & No & \textit{timeXplain}\\ 
            ROCKET-SHAP & Post-hoc & No & Local & No & \textit{timeXplain}\\ 
            MrSEQL-SM & Intrinsic & Yes & Global & Yes & \textit{MrSEQL}\\ 
            RidgeCV-SM & Intrinsic & Yes & Global & No & \textit{sklearn}\\ 
    \bottomrule
    \end{tabularx}
    \caption{Summary of properties of Explanation Methods.}
    \vspace{-12pt}
    \label{tab:explanation-method}
\end{table}


\subsection{Evaluation for Synthetic Data with Known Ground Truth}


\subsubsection{Data}

\begin{figure}[bt!]
  \centering
  \includegraphics[width=1\textwidth]{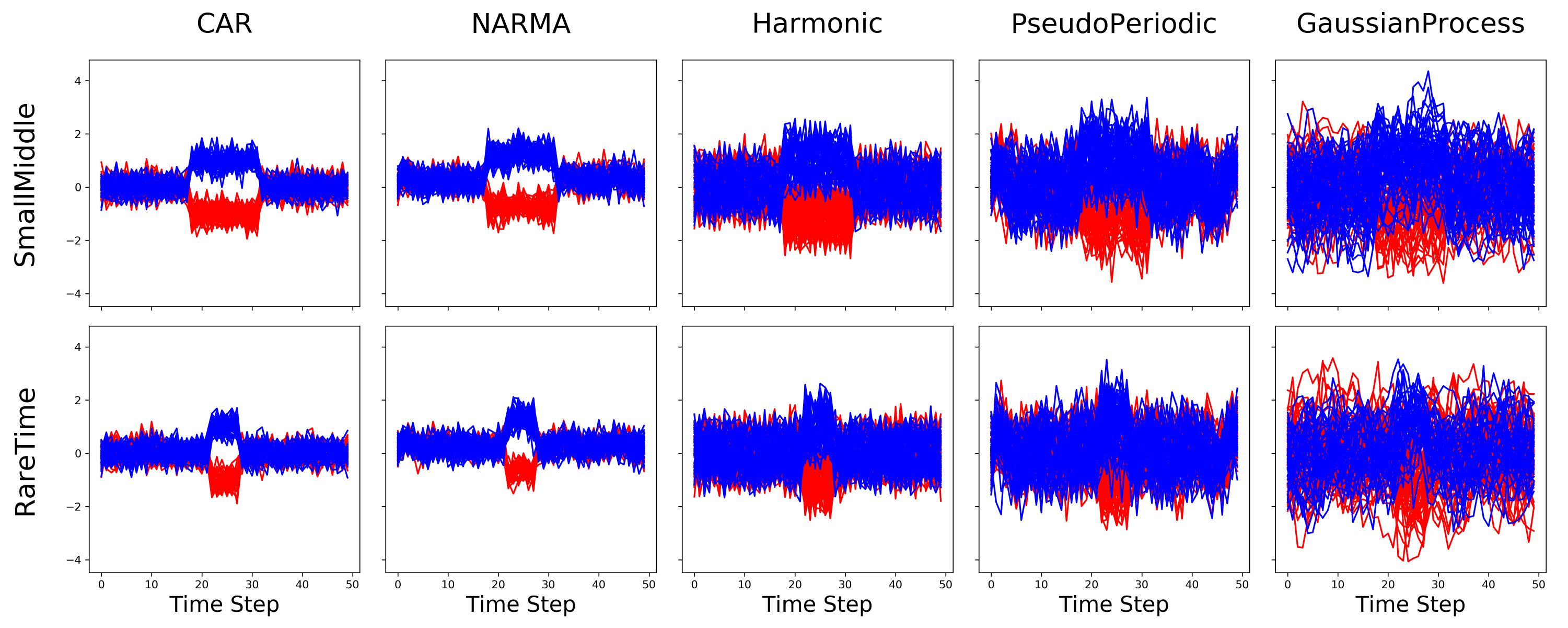}
  \caption{Visualization of the Synthetic Datasets. The columns describe the process used to create the dataset, while the row specifies the specific salient areas.}
  \vspace{5pt}
  \label{fig:synthetic}
\end{figure}
 We work with 10 synthetic univariate time series classification datasets selected by taking the mid-channel from the time series benchmark generated by \cite{Ismail2020BenchmarkingPredictions}. The datasets are created using five processes: (a) a standard continuous autoregressive time series with Gaussian noise (CAR), (b) sequences
of standard non–linear autoregressive moving average (NARMA) time series with Gaussian noise, (c) non-uniformly sampled from a harmonic function (Harmonic), (d) non-uniformly sampled from a pseudo periodic function with Gaussian noise (Pseudo Periodic), and (e) Gaussian with zero mean and
unit variance (Gaussian Process). The important areas, either a Small Middle part (30\% of time series length) or a very small part, Rare Time (10\% of time series length), are created by adding or subtracting a constant $\mu$ ($\mu$ = 1) for the positive and negative class. The number of time steps is $T$ = 50. Each dataset comprises of 500 samples in training set and 100 samples in testing set.  Figure \ref{fig:synthetic} visualizes the two classes in the 10 datasets.

Before presenting the experiment result of our evaluation on the synthetic datasets, we discuss the effect of the Data Perturbation strategy (Section \ref{subsec:impact-data-perturbation-strategy}), investigate the impact of Referees (Section \ref{subsec:impact-referees}), and perform a sanity check for the classifier quality used for model-agnostic post-hoc explanation methods such as LIME and SHAP (Section \ref{subsec:sanity-check-base-model}). 

\subsubsection{Impact of Data Perturbation Strategy}
\label{subsec:impact-data-perturbation-strategy}

\begin{figure}[!hbt]
\hspace*{-1cm}
\centering
  \includegraphics[scale=0.32]{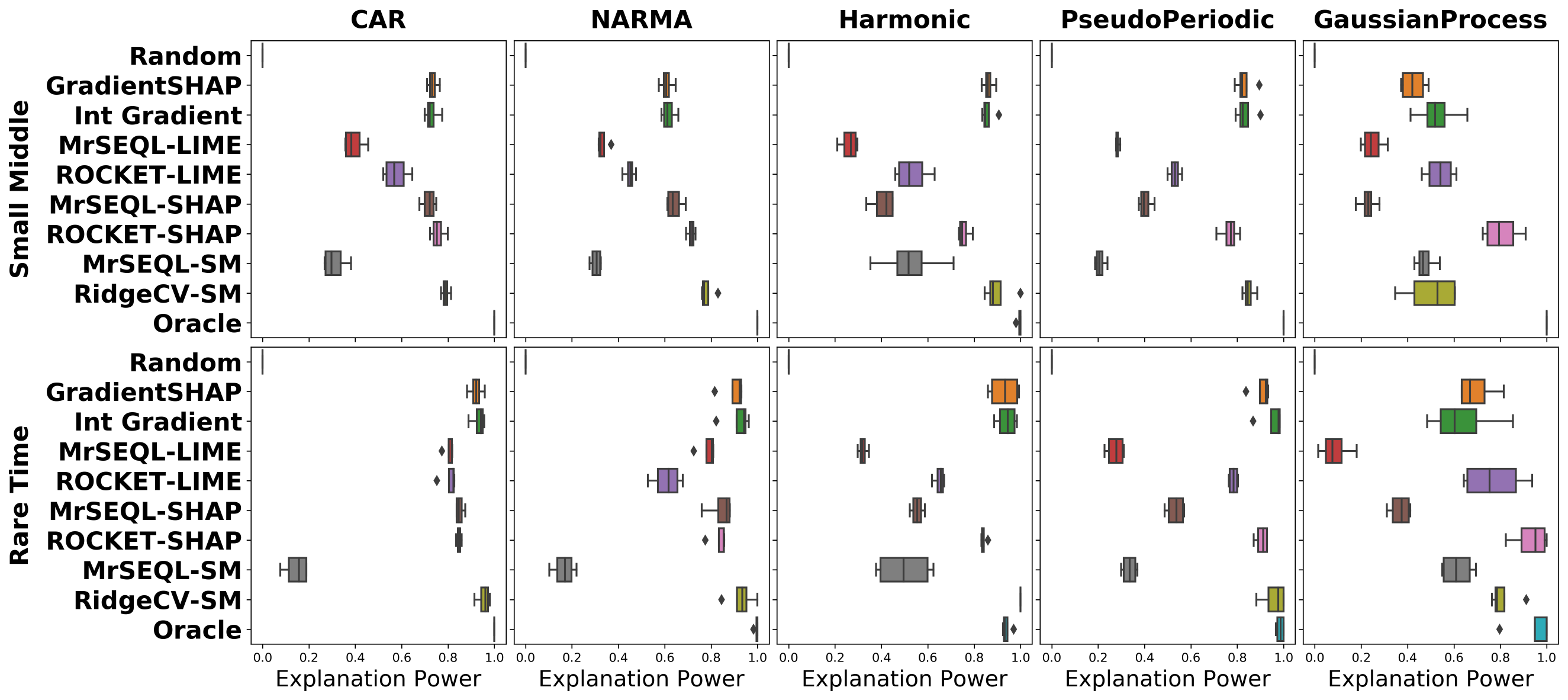}
  \caption{Impact of data perturbation strategy on  Explanation Power for each explanation method.
  A smaller box-range (which comes from 4 perturbation methods) indicates a smaller change of the Explanation Power with different perturbation strategies. 
  For datasets that are "easier" to classify by referees, this range is often smaller than that of "harder"-to-classify datasets.
  }
  \label{fig:ablation-bymasking-sup}
\end{figure}

\begin{figure}[!bt]
\hspace*{-1cm}
\centering
  \includegraphics[width=1.2\textwidth]{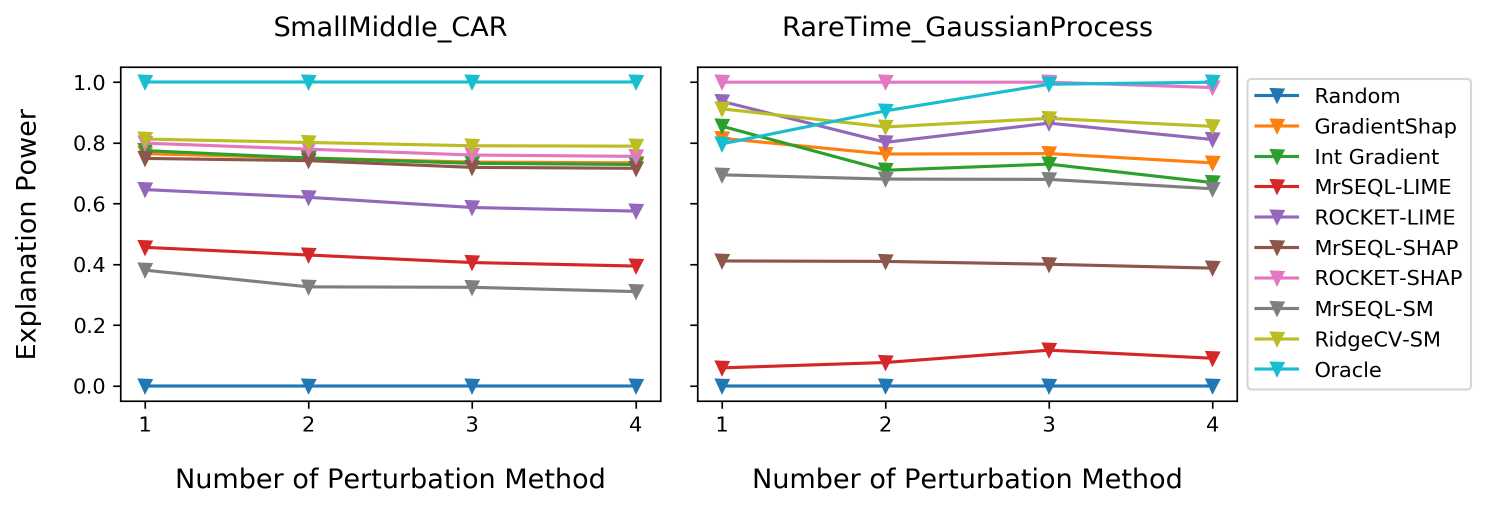}
  
  \caption{Changes of Explanation Power when different Perturbations are sequentially introduced. The sequence of perturbations is:  Local Mean, Local Gaussian, Global Mean, and Global Gaussian (less to more extreme perturbation). The two example datasets are SmallMiddle\_CAR (easy-to-classify dataset) and RareTime\_GaussianProcess (hard-to-classify dataset). For the harder dataset RareTime\_GaussianProcess, the relative position of the Explanation Methods changes, indicating that having multiple types of perturbation methods is helpful when the dataset is hard to classify. Specifically, for Oracle explanation, if only a single perturbation method was used, such as Local Mean or Local Gaussian, the evaluation result would rank Oracle explanation as the 6\ts{th} best explanation method. However, when more perturbations are introduced, Oracle explanation is evaluated more robustly, placing this method to the top 1 best explanation and closer to the ground truth.}
  
  \label{fig:ablation-bynoise-sequential}
\end{figure}

Figure \ref{fig:ablation-bymasking-sup} shows the boxplots of Explanation Power for different data perturbation strategies. In datasets which are "easier" to classify (i.e., most classifiers get close to 100\% accuracy) such as CAR and NARMA, the Explanation Power does not change with the perturbation strategy. On the other hand, we observe a larger change in Explanation Power when data is harder to classify (for example, in GaussianProcess datasets). 
We additionally present plots showing changes of Explanation Power in two extreme cases for the SmallMiddle\_CAR dataset (easy-to-classify) and RareTime\_GaussianProcess dataset (hard-to-classify) when different perturbations are gradually introduced (Figure \ref{fig:ablation-bynoise-sequential}). Notably, for the harder  dataset RareTime\_GaussianProcess, having more perturbation methods encourages the evaluation results to get closer to the ground truth. 
Specifically, for the Oracle explanation, if only a single perturbation method was used, such as Local Mean or Local Gaussian, the evaluation result would rank the Oracle explanation as the 6\ts{th} best explanation method. However, when more perturbations are introduced, the Oracle explanation is evaluated more robustly, placing this method to the top 1 best explanation and better aligned with the ground truth.
 When the explanation is Oracle (upper bound of explanation) and Random (lower bound of explanation), we generally observe that these explanations are the most and least informative informative methods, respectively.

\subsubsection{Impact of Referee Classifiers} 
\label{subsec:impact-referees}
\begin{figure}[!bt]
\hspace*{-1cm}
\centering
  \includegraphics[scale=0.32]{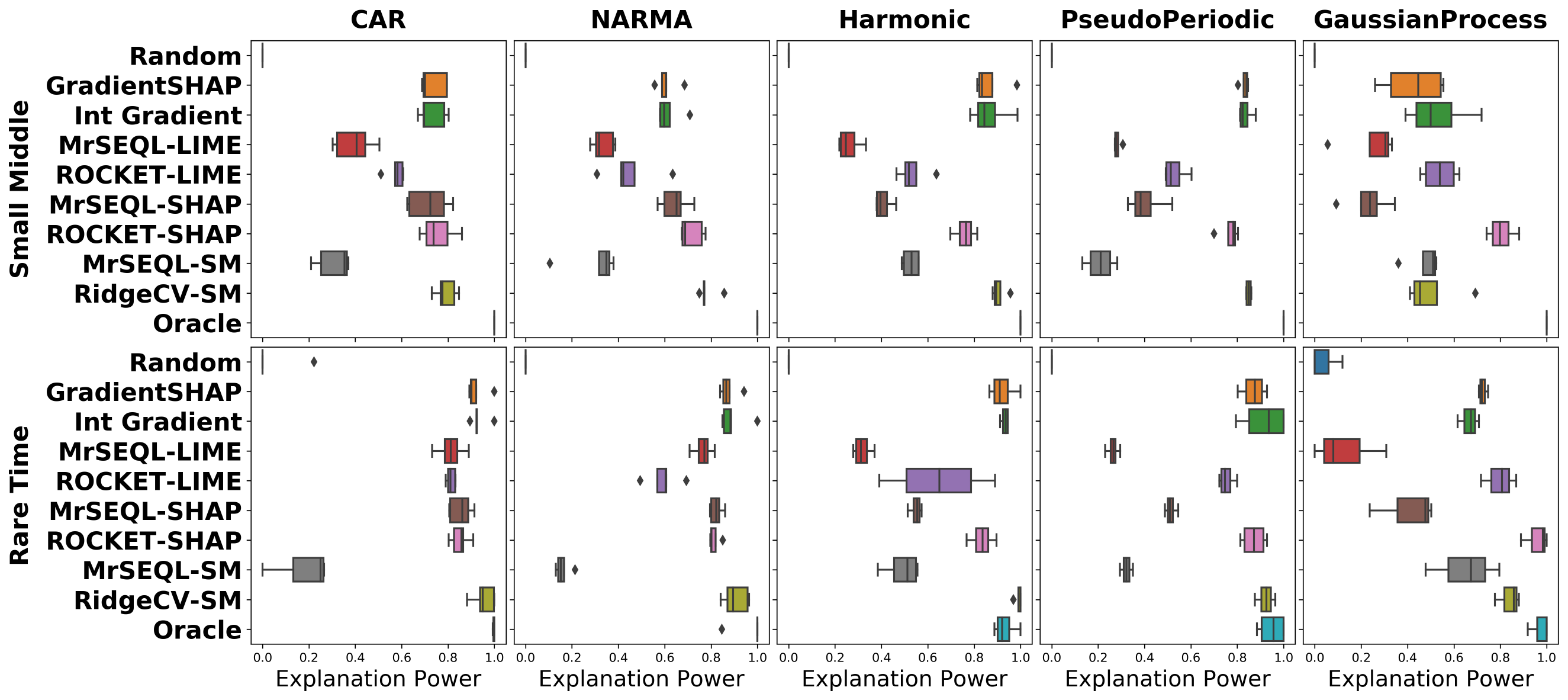}
  \caption{Impact of referees on explanation power for each explanation method. 
  A smaller box-range (which comes from 5 referee classifiers)  signals a smaller change of explanation power by different referees. 
  Besides, in a specific dataset, the relative position of this range indicates the level of critical difference in opinions of referees in their votes. 
  Nevertheless, having a committee of referees that are highly accurate is generally desirable.
  }
  \label{fig:ablation-byref-sup}
\end{figure}

Similar to the previous investigation on the impact of the perturbation  strategy, we now inspect how the Explanation Power changes with respect to the set of referees, and present the result in Figure \ref{fig:ablation-byref-sup}. Here, we also notice a relatively consistent explanation power among different referee classifiers in datasets that are easier to classify (such as CAR and NARMA datasets). In datasets that are harder-to-classify (for example, in Gaussian Process datasets), we observe a larger range in distribution of explanation methods over referee classifiers. 

\begin{figure}[!bt]
\hspace*{-1cm}
\centering
  \includegraphics[width=1.2\textwidth]{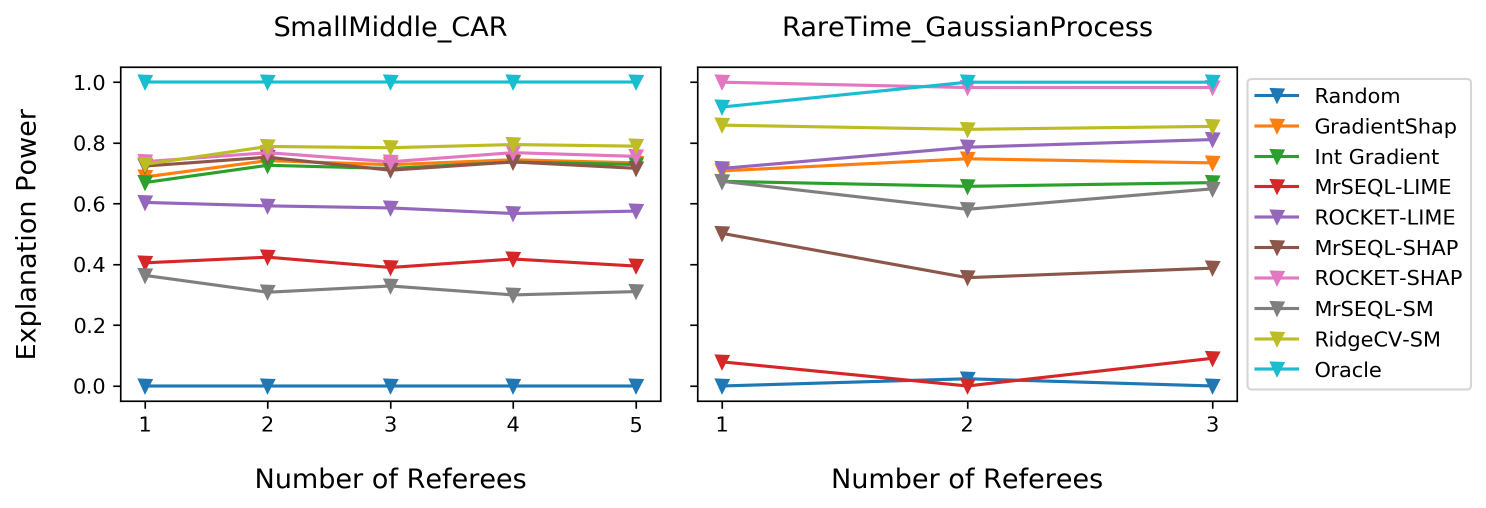}
  
  \caption{Changes of Explanation Power when different Referees are sequentially introduced. The two example datasets are SmallMiddle\_CAR (easy-to-classify
dataset) and RareTime\_GaussianProcess (hard-to-classify dataset. The sequence of addition of referees is from lowest to highest accuracy, filtered to represent the most reliable classifiers among the ones used for evaluation. Details of these sequence are in Section \ref{sec:appendix}. For the harder dataset RareTime\_ GaussianProcess, the relative position of the
Explanation Methods changes, indicating that having a set of referees is helpful and leads to more stable results.
Specifically, for Oracle explanation, if only a single referee is employed, the evaluation result could have ranked Oracle explanation as the 2\ts{nd} best explanation method. However, when more referees are introduced, Oracle explanation is evaluated more robustly, placing this method to the top 1 best explanation and closer to the ground truth.}
  
  \label{fig:ablation-byrefs-sequential}
\end{figure}

Random and Oracle explanations both have their Explanation Power in expected values for the evaluated datasets. We present the change of Explanation Power when different referees are sequentially introduced in the two extreme cases on SmallMiddle\_CAR dataset (easy-to-classify) and RareTime\_GaussianProcess dataset (hard-to-classify)(Figure \ref{fig:ablation-byrefs-sequential}). We observe that for RareTime\_GaussianProcess dataset which is hard to classify, having a committee of referees that are highly accurate is desirable and is helpful in reducing the potential bias of a single referee and can lead to a more stable evaluation. Specifically, for Oracle explanation, if only a single referee was employed, the evaluation result would  have ranked Oracle explanation as the 2\ts{nd} best explanation method. However, when more referees are introduced, Oracle explanation is evaluated more robustly, placing this method to the top 1 best explanation and closer to the ground truth.

Similarly, we note that for some real datasets, several referee classifiers that are highly accurate can disagree in their evaluation ranking. In such cases, having multiple referees leads to a considerably more robust and reliable result. We show an example using a real dataset with domain expert ground truth (the Counter Movement Jump dataset) in a later section (Section \ref{subsec:CMJ}).

\subsubsection{Sanity Check for the Impact of the Base Classifier Quality}
\label{subsec:sanity-check-base-model}
\begin{figure}[!bt]
\hspace*{-1cm}
\centering
  \includegraphics[scale=0.32]{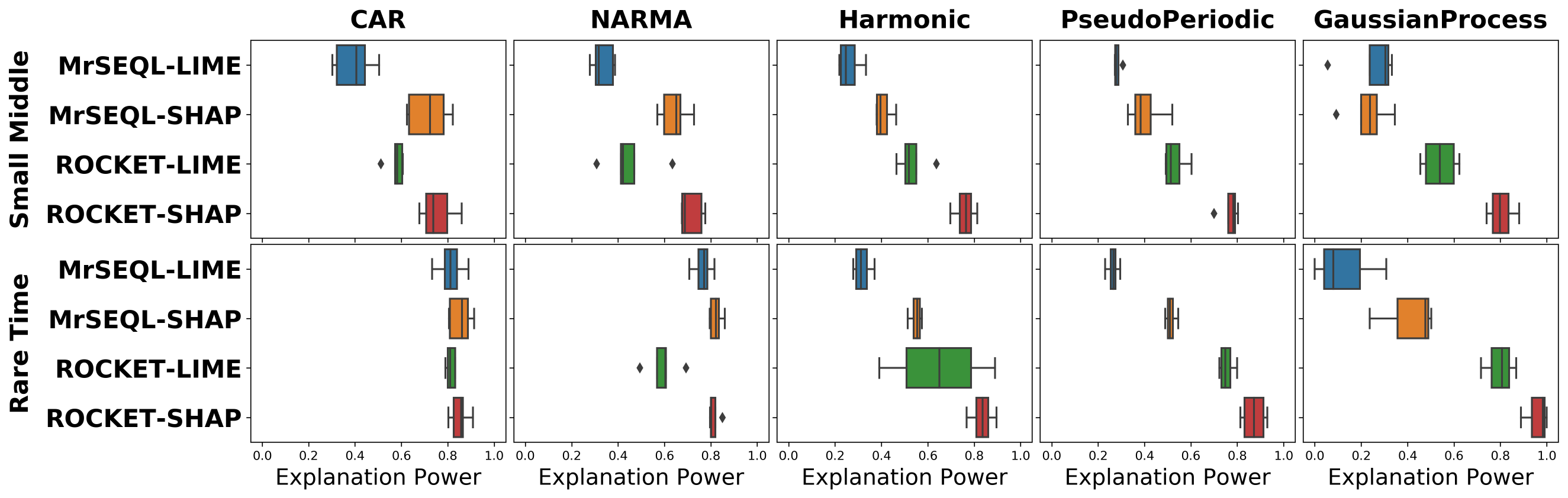}
  \caption{Sanity Check of Base Model Quality with LIME and SHAP. 
  Higher accuracy of the classifier base-model (ROCKET $>$ MrSEQL), leads to higher explanation power for the corresponding explanations (ROCKET-LIME and ROCKET-SHAP have higher explanation power than MrSEQL-LIME and MrSEQL-SHAP).
  }
  \label{fig:sanity-check}
\end{figure}

Model-agnostic post-hoc methods such as LIME and SHAP derive explanations based on a classifier of any type. Thus, these explanation are dependent on the performance of the base classifier. For example, the ROCKET-SHAP explanation is created by applying SHAP (explanation method) on ROCKET (base classifier).
If the base classifier has low accuracy on the sample dataset, the explanation based on that classifier may not be as good as one based on a more accurate classifier. In our experiment, we get LIME and SHAP explanations from two sources: MrSEQL classifier \cite{LeNguyen2019InterpretableRepresentations} and ROCKET classifier  \cite{Dempster2020ROCKET:Kernels}. 
We observe that ROCKET achieves higher accuracy than MrSEQL in datasets created from Pseudo Periodic, Harmonic, and Gaussian Process (Table \ref{tab:acc_synthetic} in Appendix). We compare the two pairs of explanation (MrSEQL-based and ROCKET-based) from LIME and SHAP and do a sanity check. Our experiment confirms that in both cases, under the AMEE evaluation approach, ROCKET-LIME and ROCKET-SHAP are considered better explanation methods as  compared to MrSEQL-LIME and MrSEQL-SHAP, respectively (Figure \ref{fig:sanity-check}). This sanity check confirms our intuition that the quality of the base classifier is an important factor in model-agnostic, post-hoc explanation methods such as LIME and SHAP. 

\subsubsection{Results}
Using a committee of 5 referee classifiers and 4 data perturbation strategies, we evaluate 10 explanation methods (8 computed explainers plus the lower bound explanation (Random) and upper bound explanation (Oracle)) using AMEE. The resulting Explanation Power is presented in  
Table \ref{tab:result_synthetic}. 

Using a threshold of 0.5 of the min-max normalized saliency score in [0,1], we determine the ground truth of whether a time point is salient. 
We compare the explanation methods with the ground truth explanation for each time points and calculate the F1-score (Table \ref{tab:result_synthetic-f1-score}) to measure how good each method is in determining saliency of a time point. 
For example, in the SmallMiddle\_CAR dataset, AMEE selects the best explanation method to be Oracle with Explanation Power of 1.00 (Table \ref{tab:result_synthetic}), and the second best explanation is the Saliency Map from RidgeCV (Explanation Power of 0.79, Table \ref{tab:result_synthetic}). This result is similar to the F1-score of these explanations using the ground truth (Table \ref{tab:result_synthetic-f1-score}). Here, the Oracle explanation achieves 1.00 in F1-score (highest), followed by RidgeCV saliency map, achieving 0.86 in F1-score (second-best). Similarly, we find that the ranks of the methods in Table \ref{tab:result_synthetic} and \ref{tab:result_synthetic-f1-score} in high agreement. Moreover, even for the hardest dataset to classify (RareTime\_GaussianProcess), we observe that adding more referees brings the  relative ranking of the evaluated explanation closer to the ground truth ranking(Table \ref{tab:more-referee}). 
This result further reinforces that using  multiple referees is desirable: we observe that a highly accurate set of referees brings the explanation ranking closer to the ground truth.
Overall, our result shows a high agreement between AMEE's computed Explanation Power and the F1-score using ground truth time-point importance evaluation and confirms that the committee of referees is a desirable property of the explainer recommendation framework.

 \setlength{\tabcolsep}{2pt}
 \renewcommand{\arraystretch}{1.1}
\begin{table}[!hbt]
    \centering
    \scalebox{0.75}{%
    \begin{tabular}{|l||c c c c c c c c c c|}
    \hline
   
        Dataset & Random & \shortstack{Grad\\SHAP} & 
        \shortstack{Int\\Gradient} & \shortstack{MRSEQL\\-LIME} & \shortstack{ROCKET\\-LIME} & \shortstack{MRSEQL\\-SHAP} & \shortstack{ROCKET\\-SHAP} & \shortstack{MRSEQL\\-SM} & \shortstack{RIDGECV\\-SM} & 
        \shortstack{Oracle} \\ \hline
       
        SM\_CAR & 0.00 & 0.73 & 0.73 & 0.39 & 0.58 & 0.72 & 0.76 & 0.31 & 0.79 & 1.00 \\ 
        SM\_NARMA & 0.00 & 0.61 & 0.62 & 0.33 & 0.45 & 0.64 & 0.71 & 0.30 & 0.78 & 1.00 \\ 
        SM\_Harmonic & 0.00 & 0.87 & 0.86 & 0.26 & 0.54 & 0.41 & 0.76 & 0.53 & 0.91 & 1.00 \\ 
        SM\_PseudoPeriodic & 0.00 & 0.83 & 0.84 & 0.28 & 0.53 & 0.40 & 0.77 & 0.21 & 0.85 & 1.00 \\ 
        SM\_GaussianProcess & 0.00 & 0.43 & 0.53 & 0.25 & 0.54 & 0.23 & 0.80 & 0.48 & 0.50 & 1.00 \\ \hline
        
        RT\_CAR & 0.00 & 0.92 & 0.93 & 0.81 & 0.81 & 0.85 & 0.85 & 0.14 & 0.95 & 1.00 \\ 
        RT\_NARMA & 0.00 & 0.90 & 0.92 & 0.79 & 0.61 & 0.85 & 0.84 & 0.17 & 0.93 & 1.00 \\ 
        RT\_Harmonic & 0.00 & 0.93 & 0.94 & 0.32 & 0.65 & 0.55 & 0.84 & 0.50 & 1.00 & 0.94 \\ 
        RT\_PseudoPeriodic & 0.00 & 0.92 & 0.96 & 0.28 & 0.79 & 0.54 & 0.92 & 0.34 & 0.97 & 1.00 \\ 
        RT\_GaussianProcess & 0.00 & 0.73 & 0.67 & 0.09 & 0.81 & 0.39 & 0.98 & 0.65 & 0.85 & 1.00 \\ 
    \hline        
    \end{tabular}
    }
    \smallskip
    \caption{Synthetic Datasets: Explanation Power for each of the 10 explanation methods evaluated.} \label{tab:result_synthetic}
    
\end{table}

 \setlength{\tabcolsep}{2pt}
  \renewcommand{\arraystretch}{1.1}
\begin{table}[!hbtp]
    \centering
    \scalebox{0.75}{%
    \begin{tabular}{|l||c c c c c c c c c c|}
    \hline
        Dataset & Random & \shortstack{Grad\\SHAP} & 
        \shortstack{Int\\Gradient} & \shortstack{MRSEQL\\-LIME} & \shortstack{ROCKET\\-LIME} & \shortstack{MRSEQL\\-SHAP} & \shortstack{ROCKET\\-SHAP} & \shortstack{MRSEQL\\-SM} & \shortstack{RIDGECV\\-SM} & 
        \shortstack{Oracle} \\ \hline
        SM\_CAR & 0.36 & 0.81 & 0.83 & 0.59 & 0.70 & 0.83 & 0.83 & 0.50 & 0.86 & 1.00 \\ 
        SM\_NARMA & 0.36 & 0.75 & 0.77 & 0.61 & 0.66 & 0.83 & 0.83 & 0.43 & 0.84 & 1.00 \\ 
        SM\_Harmonic & 0.36 & 0.56 & 0.59 & 0.51 & 0.52 & 0.63 & 0.80 & 0.44 & 0.61 & 1.00 \\ 
        SM\_PseudoPeriodic & 0.36 & 0.54 & 0.56 & 0.45 & 0.58 & 0.58 & 0.74 & 0.55 & 0.56 & 1.00 \\ 
        SM\_GaussianProcess & 0.36 & 0.19 & 0.24 & 0.42 & 0.50 & 0.38 & 0.65 & 0.61 & 0.16 & 1.00 \\ \hline
        RT\_CAR & 0.21 & 0.84 & 0.87 & 0.59 & 0.64 & 0.74 & 0.72 & 0.16 & 0.92 & 1.00 \\ 
        RT\_NARMA & 0.21 & 0.78 & 0.85 & 0.60 & 0.50 & 0.72 & 0.74 & 0.16 & 0.91 & 1.00 \\ 
        RT\_Harmonic & 0.21 & 0.42 & 0.47 & 0.29 & 0.39 & 0.46 & 0.69 & 0.29 & 0.65 & 1.00 \\ 
        RT\_PseudoPeriodic & 0.21 & 0.51 & 0.55 & 0.25 & 0.47 & 0.37 & 0.63 & 0.29 & 0.68 & 1.00 \\ 
        RT\_GaussianProcess & 0.21 & 0.21 & 0.20 & 0.22 & 0.37 & 0.25 & 0.51 & 0.34 & 0.33 & 1.00 \\ \hline
    \end{tabular}
    }
    \smallskip
    \caption{Synthetic Datasets: F1-score of explanation methods using explanation ground truth.
    \textit{Abbreviation:} SM - Small Middle, RT - Rare Time.
    }
    
    \label{tab:result_synthetic-f1-score}
\end{table}

\newcolumntype{Y}{>{\centering\arraybackslash}X}

\begin{table}[t]
\centering
\begin{tabularx}{.9\textwidth}{l *{3}{Y} *{1}{Y} }

\toprule
                     & \multicolumn{3}{c}{Rank by Number of Referees} 
                     &           
                      \multicolumn{1}{c}{Rank of F1}                 \\ 
                     \cmidrule(lr){2-4} 
                     
Explanation Method   & \textbf{1}                  & \textbf{2}                 & \textbf{3}                              & GroundTruth    \\ 
\midrule
\textbf{Oracle}        & \textbf{2}   & \textbf{1}                    & \textbf{1}  & \textbf{1}                         \\
\textbf{ROCKET-SHAP}  & \textbf{1} &\textbf{2}& \textbf{2}                   & \textbf{2}                     \\     
RidgeCV-SM          & 3                    & 3                                    & 3      & 5                    \\
ROCKET-LIME          & 4                    & 4 &4&3                   \\

\textbf{GradientSHAP}          & \textbf{4}                    & \textbf{5}                                 & \textbf{5}                    & \textbf{7}                                     \\
\bottomrule
\end{tabularx}

\smallskip
    \caption{Dataset RareTime\_Gaussian Process: Explanation Power rank when referees are added sequentially for Top 5 Explanation methods. The sequence of referees is determined by the order of accuracy (least to most accurate, see  Table \ref{tab:acc_synthetic}). With the introduction of more referees, the rank of Explanation Power gets closer to the rank using F1 Ground Truth.}
    
    \label{tab:more-referee}
\end{table}

\subsubsection{Comparison with Previous Work}

Previous work \cite{Nguyen2020AClassification} showcases initial results towards comparing Explanation Methods. However, the method utilizes only one type of perturbation which  replaces salient areas with Gaussian noise of low magnitude. 
While the magnitude of the Gaussian noise can be customized, it requires extra work from users to determine this parameter.
This initial work also does not propose a way to standardise the explanation AUC across methods and datasets. 
Our new framework employs a combination of perturbation types that allows a higher impact in changing the original signals, resulting in a more robust framework and better results. We include the results of the framework (using the default settings) introduced in \cite{Nguyen2020AClassification} in Table \ref{tab:result_synthetic_prevwork}. 

This table shows that for many of the synthetic datasets, the small perturbation (default setting) added into the signal is too little and it fails to trigger changes in the classification accuracy. The resulting outcome is that the past approach is unable to distinguish differences in the informativeness of different explanation methods.

 \setlength{\tabcolsep}{2pt}
 \renewcommand{\arraystretch}{1.1}
\begin{table}[!hbt]
    \centering
    \scalebox{0.75}{%
    \begin{tabular}{|l||c c c c c c c c c c|}
    \hline
   
        Dataset & Random & \shortstack{Grad\\SHAP} & 
        \shortstack{Int\\Gradient} & \shortstack{MRSEQL\\-LIME} & \shortstack{ROCKET\\-LIME} & \shortstack{MRSEQL\\-SHAP} & \shortstack{ROCKET\\-SHAP} & \shortstack{MRSEQL\\-SM} & \shortstack{RIDGECV\\-SM} & 
        \shortstack{Oracle} \\ \hline
       
        SM\_CAR             & 0.00       & 0.00 & 0.00 & 0.00         & 0.00         & 0.00        & 0.00        & 0.00   & 0.00    & 0.00 \\
SM\_GaussianProcess & 0.42       & 0.79 & 0.15 & 0.70         & 0.48         & 1.00        & 0.60        & 0.78   & 0.54    & 0.00 \\
SM\_Harmonic        & 0.00       & 0.00 & 0.00 & 0.00         & 0.00         & 0.00        & 0.00        & 0.00   & 0.00    & 0.00 \\
SM\_NARMA           & 0.00       & 0.00 & 0.00 & 0.00         & 0.00         & 0.00        & 0.00        & 0.00   & 0.00    & 0.00 \\
SM\_PseudoPeriodic  & 0.00       & 0.00 & 0.00 & 0.00         & 0.00         & 0.00        & 0.00        & 0.00   & 0.00    & 0.00 \\ \hline
        
        RT\_CAR                & 1.00       & 0.00 & 0.00 & 0.00         & 0.00         & 0.00        & 0.00        & 0.00   & 1.00    & 0.00 \\
RT\_GaussianProcess    & 0.31       & 0.09 & 0.07 & 0.39         & 0.58         & 0.00        & 0.66        & 0.51   & 0.55    & 1.00 \\
RT\_Harmonic           & 0.52       & 0.38 & 0.36 & 0.00         & 0.74         & 0.50        & 1.00        & 0.54   & 0.69    & 0.35 \\
RT\_NARMA              & 0.00       & 1.00 & 1.00 & 0.00         & 1.00         & 0.00        & 1.00        & 0.00   & 0.00    & 0.00 \\
RT\_PseudoPeriodic     & 0.95       & 0.69 & 0.57 & 0.55         & 0.00         & 0.31        & 0.11        & 0.48   & 0.36    & 1.00 \\ 
    \hline        
    \end{tabular}
    }
    \smallskip
    \caption{Synthetic Datasets: Using default perturbation settings with previous work in \cite{Nguyen2020AClassification} to get Explanation Power for each of the 10 explanation methods evaluated.} 
    \label{tab:result_synthetic_prevwork}
    
\end{table}

\begin{figure}[!htb]
\begin{subfigure}{.5\textwidth}
  \centering
  \includegraphics[width=\linewidth]{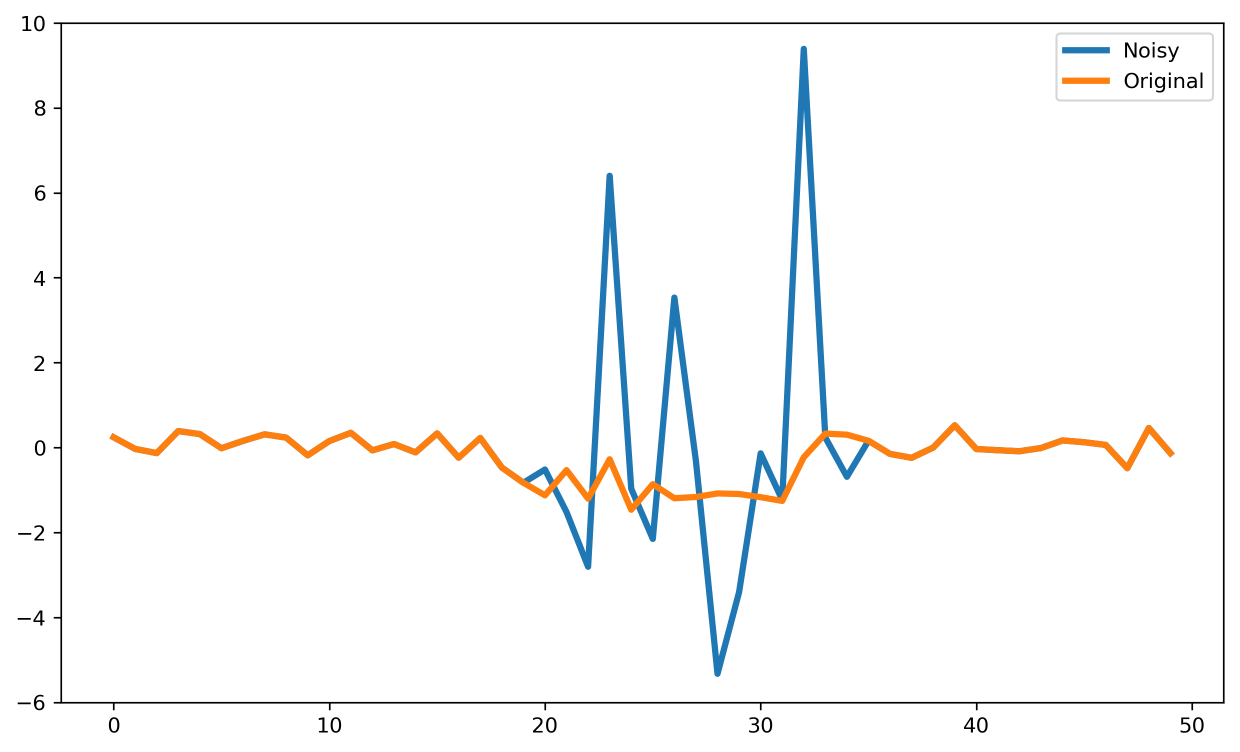}  
  \caption{Extreme Gaussian Noise addition (from \cite{Nguyen2020AClassification})}
\end{subfigure}
\begin{subfigure}{.5\textwidth}
  \centering
  \includegraphics[width=\linewidth]{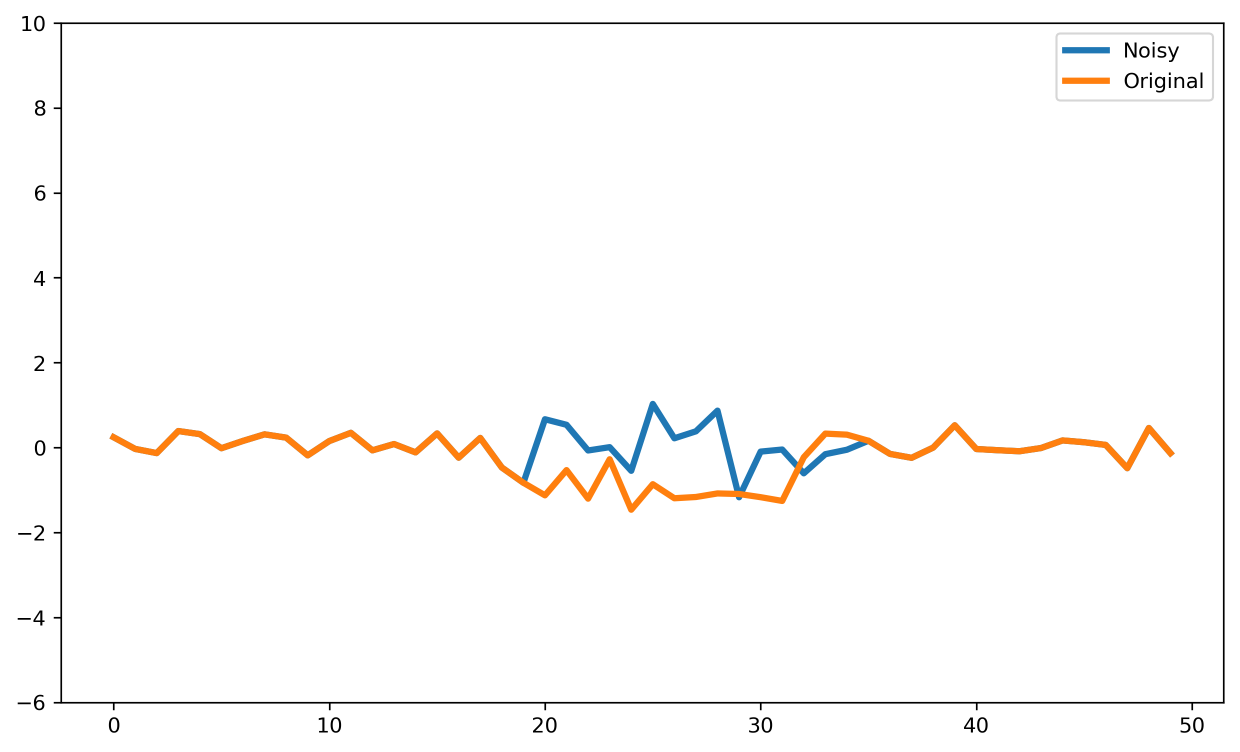}  
  \caption{Global Gaussian Noise (AMEE)
  \newline
  }
\end{subfigure}
\caption{Sample time series from dataset SmallMiddle\_CAR that was perturbed by (a) Gaussian noise addition proposed in \cite{Nguyen2020AClassification} at very high magnitude (left) and (b) Global Gaussian noise - a non-parametric perturbation) (right)}
\label{fig:compare-noise}
\end{figure}

Even with a much larger noise level (Figure \ref{fig:compare-noise}), the previous framework does not provide a result as accurate as AMEE, especially for datasets that are difficult to classify. We include results of the framework introduced in \cite{Nguyen2020AClassification} using higher noise magnitude in Table \ref{tab:result_synthetic_prevwork_maxrange} . 
\

 \setlength{\tabcolsep}{2pt}
 \renewcommand{\arraystretch}{1.1}
\begin{table}[!hbt]
    \centering
    \scalebox{0.75}{%
    \begin{tabular}{|l||c c c c c c c c c c|}
    \hline
   
        Dataset & Random & \shortstack{Grad\\SHAP} & 
        \shortstack{Int\\Gradient} & \shortstack{MRSEQL\\-LIME} & \shortstack{ROCKET\\-LIME} & \shortstack{MRSEQL\\-SHAP} & \shortstack{ROCKET\\-SHAP} & \shortstack{MRSEQL\\-SM} & \shortstack{RIDGECV\\-SM} & 
        \shortstack{Oracle} \\ \hline
       
        SM\_CAR             & 0.00       & 0.29 & 0.31 & 0.43         & 0.29         & 0.43        & 0.44        & 0.18   & 0.49    & 1.00 \\
SM\_NARMA           & 0.00       & 0.26 & 0.23 & 0.09         & 0.04         & 0.21        & 0.26        & 0.27   & 0.27    & 1.00 \\
SM\_Harmonic        & 0.30       & 0.61 & 0.55 & 0.15         & 0.67         & 0.32        & 0.44        & 0.00   & 0.60    & 1.00 \\
SM\_PseudoPeriodic  & 0.06       & 0.66 & 0.61 & 0.67         & 0.38         & 0.30        & 0.61        & 0.00   & 0.74    & 1.00 \\
SM\_GaussianProcess & 0.00       & 0.32 & 0.30 & 0.34         & 0.57         & 0.22        & 0.30        & 0.63   & 0.09    & 1.00 \\ \hline
        
        RT\_CAR             & 0.00       & 0.78 & 0.86 & 0.80         & 0.59         & 0.82        & 0.70        & 0.25   & 0.81    & 1.00 \\
RT\_NARMA           & 0.17       & 0.54 & 0.51 & 0.15         & 0.00         & 0.25        & 0.17        & 0.24   & 0.46    & 1.00 \\
RT\_Harmonic        & 0.66       & 0.64 & 0.68 & 0.40         & 0.86         & 0.70        & 1.00        & 0.00   & 0.62    & 0.80 \\
RT\_PseudoPeriodic  & 0.00       & 0.60 & 0.68 & 1.00         & 0.65         & 0.27        & 0.49        & 0.19   & 0.25    & 0.94 \\
RT\_GaussianProcess & 0.19       & 0.00 & 0.12 & 0.37         & 0.91         & 0.59        & 1.00        & 0.53   & 0.59    & 0.48 \\ 
    \hline        
    \end{tabular}
    }
    \smallskip
    \caption{Synthetic Datasets: Using higher perturbation magnitude with previous work in \cite{Nguyen2020AClassification} to get Explanation Power for each of the 10 explanation methods evaluated. Even with extreme perturbation, results of previous work did not agree with the F1-measure in Table \ref{tab:result_synthetic-f1-score} as AMEE's Explanation Power does.} 
    \label{tab:result_synthetic_prevwork_maxrange}
    
\end{table}

\subsection{Evaluation for Real  Time Series Data}
\label{sec:exp:real}
\subsubsection{Data}
We work with 15 datasets from the UCR Archive  \cite{Dau2018TheArchive} that represent a variety of data sources and domains.
These datasets are of 5 types: 
electrocardiogram (ECG), human motion (MOTION), device usage (DEVICE), device   activities tracked by sensors (SENSOR) and spectroscopy (SPECTRO). 
Oracle explanation is not available for these datasets.

\subsubsection{Results} We test explanations for these datasets with AMEE and report the result in Table \ref{tab:result_ucr}. Since we do not have ground truth for the majority of these datasets, we use this experiment to show how AMEE can apply to real datasets. We note that the Random explanation sometimes outperforms a method-base explanation. This can happen as some explanation methods may not work well with certain datasets, resulting in \textit{unreasonable} explanations that misleadingly highlight non-discriminative parts as discriminative, or fail to identify any significant discriminative parts at all. In this situation, the evaluation of random explanations can serve as a filter for \textit{reasonable} explanation methods, and any methods that have lower performance than random should be filtered out.

 \setlength{\tabcolsep}{2pt}
  \renewcommand{\arraystretch}{1.1}

\begin{table}[!tb]
    \hspace*{-0.5cm}
    \centering
    \scalebox{0.8}{%
    \begin{tabular}{|l|l||ccccccccc|}
    \hline
        Data Type & Dataset & Random & \shortstack{Grad\\SHAP} & 
        \shortstack{Int\\Gradient} & \shortstack{MRSEQL\\-LIME} & \shortstack{ROCKET\\-LIME} & \shortstack{MRSEQL\\-SHAP} & \shortstack{ROCKET\\-SHAP} & \shortstack{MRSEQL\\-SM} & \shortstack{RIDGECV\\-SM}  \\ \hline
        
ECG        & ECG200      & 0.00       & 0.13 & 0.17 & 0.67         & 0.57         & 0.84        & \textbf{1.00}        & 0.46   & 0.45    \\
           & ECG5000     & 0.00       & \textbf{1.00} & 0.99 & 0.67         & 0.57         & 0.61        & 0.82        & 0.30   & 0.29    \\
           & ECGFiveDays & 0.54       & 0.77 & 0.71 & 0.39         & 0.54         & 0.91        & \textbf{1.00}        & 0.00   & 0.27    \\
           & TwoLeadECG  & 0.39       & 0.17 & 0.18 & 0.44         & 0.40         & 0.92        & \textbf{1.00}        & 0.05   & 0.00    \\ \hline
MOTION     & GunPoint    & 0.81       & 0.74 & \textbf{1.00} & 0.77         & 0.00         & 0.92        & 0.84        & 0.89   & 0.56    \\
           & CMJ         & 0.24       & 0.06 & 0.00 & 0.99         & 0.56         & \textbf{1.00}        & 0.84        & 0.90   & 0.31    \\\hline
DEVICE     & PowerCons   & 0.50       & 0.69 & 0.68 & 0.37         & 0.76         & 0.45        & \textbf{1.00}        & 0.00   & 0.66    \\\hline
SPECTRO    & Coffee      & 0.00       & 0.36 & 0.53 & 0.53         & 0.37         & 0.86        & \textbf{1.00}        & 0.72   & 0.57    \\
           & Strawberry  & 0.65       & 0.41 & 0.43 & 0.91         & 0.66         & \textbf{1.00}        & 0.40        & 0.70   & 0.00    \\\hline
SENSOR     & Car         & 0.45       & 0.15 & 0.00 & 0.42         & 0.36         & \textbf{1.00}        & 0.74        & 0.33   & 0.44    \\
           & ItalyPower  & 0.27       & \textbf{1.00} & 0.95 & 0.16         & 0.13         & 0.11        & 0.29        & 0.00   & 0.54    \\
           & Plane       & 0.75       & 0.80 & 0.78 & 0.50         & 0.00         & 0.67        & \textbf{1.00}        & 0.19   & 0.32    \\
           & Sony1       & 0.23       & 0.84 & 0.80 & 0.21         & 0.52         & 0.32        & \textbf{1.00}        & 0.00   & 0.83    \\
           & Sony2       & 0.32       & 0.99 & \textbf{1.00} & 0.45         & 0.41         & 0.37        & 0.82        & 0.00   & 0.57    \\
           & Trace       & 0.33       & 0.25 & 0.21 & 0.77         & 0.26         & \textbf{1.00}        & 0.79        & 0.16   & 0.00 \\  \hline
          ~ & \textbf{Average}          & 0.37 & 0.56 & 0.56 & 0.55 & 0.41 & 0.73 & \textbf{0.84} & 0.31 & 0.39 \\
          \hline
          ~ & \textbf{Count of 1.00}          & 0/15 & 2/15  & 2/15 & 0/15 & 0/15 & 4/15 & \textbf{7/15} & 0/15 & 0/15 \\\hline
    \end{tabular}
    }
    \smallskip
    \caption{Explanation Power on UCR Datasets. Most informative method is highlighted in \textbf{bold}. The last row shows the count of occurrences when a method is selected as the most informative explanation according to AMEE.}
    
    \label{tab:result_ucr}
\end{table}

\subsection{Evaluation for Real Dataset with Expert Ground Truth}
\label{subsec:case-study}

The Oracle explanation is the upperbound for any explanation method, however, it is only available in synthetic datasets. For real datasets the explanation ground truth is often available in an approximate level of precision, e.g., specifying the relative position of the shape and areas of importance. 
This approximate ground truth is widely used in other papers in evaluating explanation methods for images \cite{Kim2018Interpretabilitytcav, Zhou2016LearningLocalization, Selvaraju2017Grad-cam:Localization}, however, this approximate ground truth is not readily available for time series data without opinions from data experts. 
Among the the datasets evaluated in Section \ref{sec:exp:real}, Coffee \cite{coffee}, Counter Movement Jump (CMJ) \cite{LeNguyen2019InterpretableRepresentations}, and GunPoint \cite{gunpoint} have this information  of the true important areas for each class of the dataset. In this section, we will compare the  saliency-based explanations evaluated by AMEE, and the expert ground truth of important areas.

\subsubsection{Spectroscopy Dataset: Coffee} 

\begin{figure}[ht!]
\vspace*{5pt}
  \centering
    \includegraphics[width=0.6\textwidth]{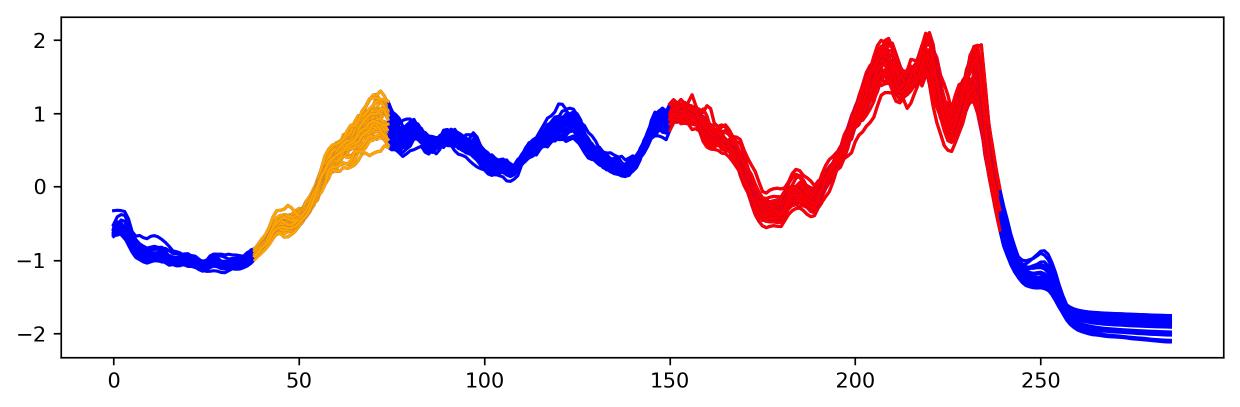}
    
  \caption{Coffee Dataset: Ground truth from Coffee dataset\cite{coffee}. According to the original paper \cite{coffee}, the chlorogenic acid content (region approximately of time steps 150-240, marked in red) is the major region that contributes to the difference in two coffee types. The caffeine content (region approximately of time steps 40-75, marked in orange) is also discriminative, but to a lesser extent.}
  \label{fig:groundtruth-coffee}
\end{figure}

\begin{figure}[!htb]
\centering
  \includegraphics[width=1\textwidth]{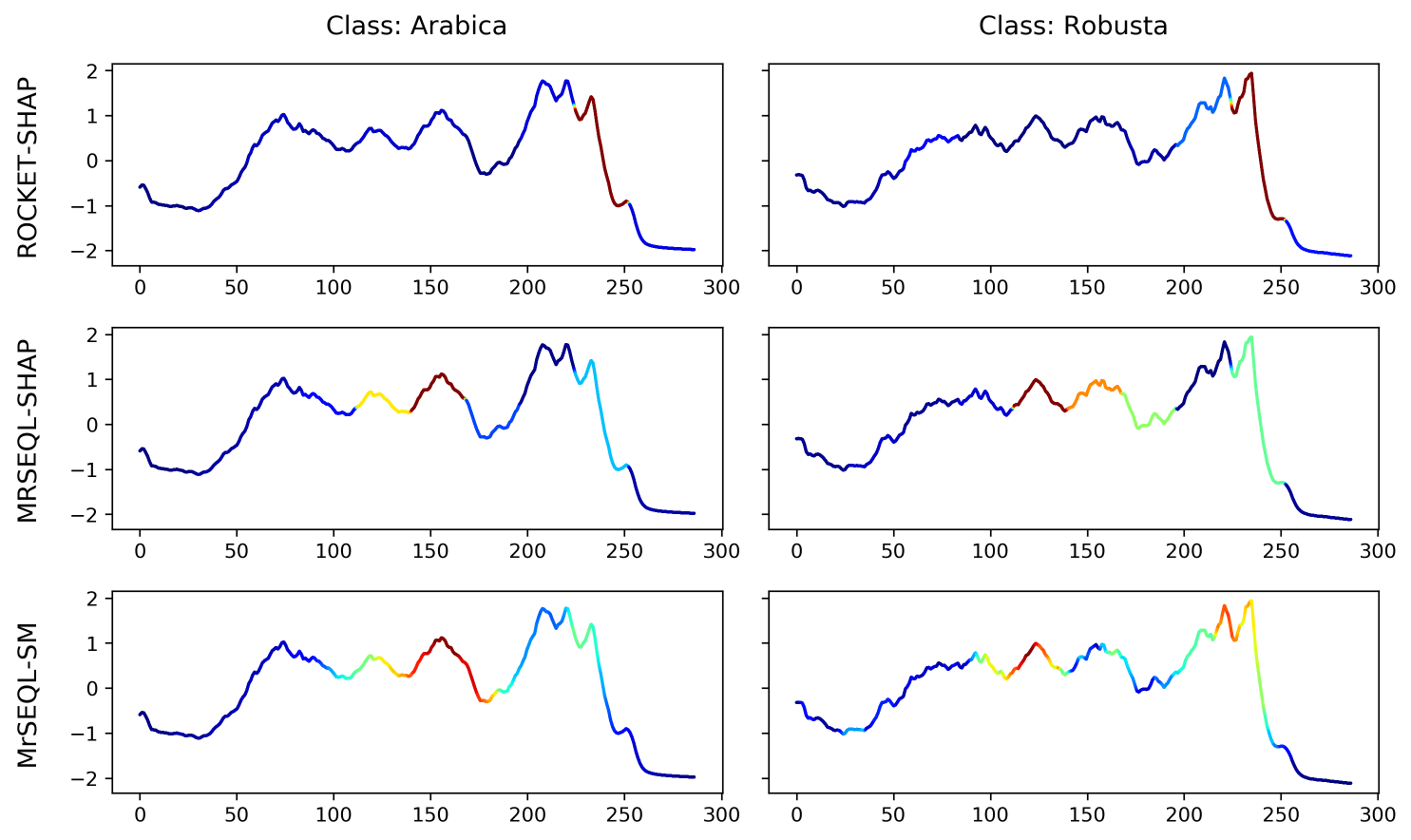}
  \rule[1ex]{11.8cm}{1pt}
  \includegraphics[width=1\textwidth]{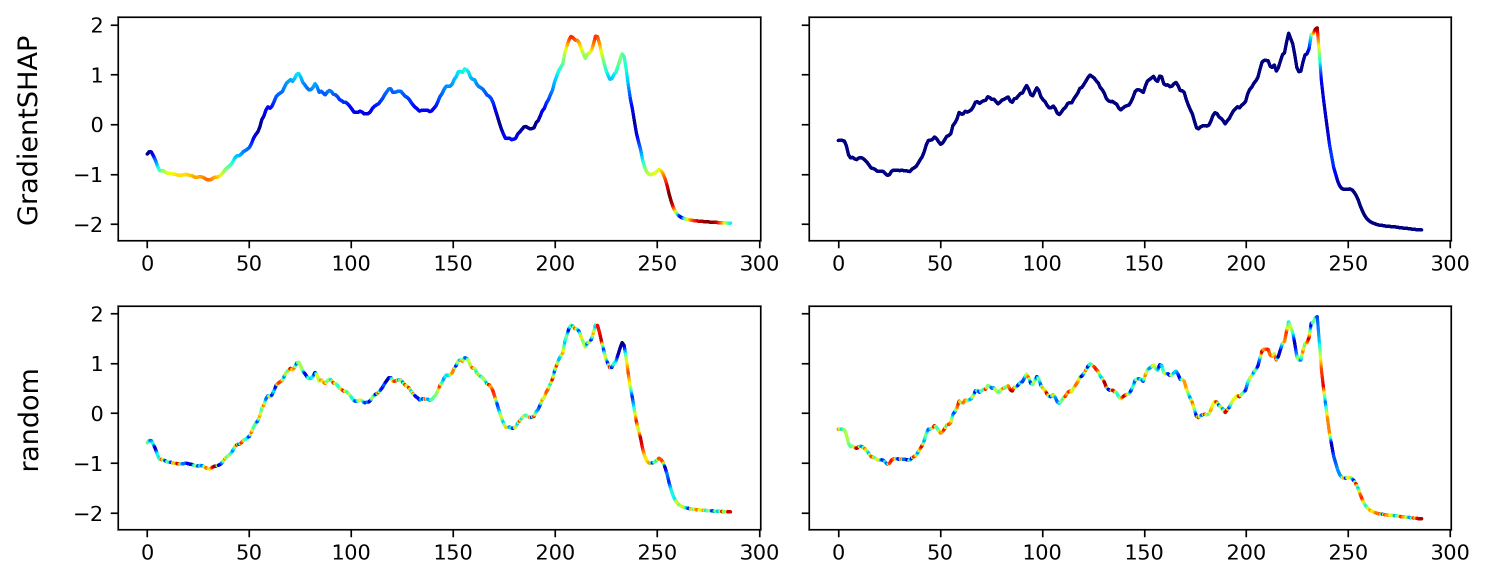}
  
  \caption{Coffee Dataset: Visualization of top 3 best explanation methods (top 3 rows) and worst explanation method (bottom 2 rows) ranked by AMEE. It can be observed that the top explanation methods are all able to point out the discriminative regions that are confirmed by the domain expert, as represented in Figure \ref{fig:groundtruth-coffee}.}
  \label{fig:case-study-coffee}
\end{figure}

The Coffee dataset contains the  spectroscopy sample of two types of coffee: Arabica and Robusta. This dataset was first introduced in \cite{coffee} and is part of the UCR time series dataset \cite{Dau2018TheArchive}. 
Figure \ref{fig:case-study-coffee} shows the top 3 and bottom 2 explanation methods ranked by AMEE. Notably, the discriminating region of the two classes of Coffee produced by the best explainer, ROCKET-SHAP, is the last peak region of the time series. This region was confirmed in the original paper \cite{coffee} to contain information about the chlorogenic acid content of the sample that contributes to the difference between the two types of coffee (Figure
\ref{fig:groundtruth-coffee}). Arabica has a lower caffeine and chlorogenic acid content that contributes to its finer taste and greater market value. The region that the MrSEQL-SHAP and MrSEQL-SM also highlight is another part of the spectrum that contains information about the chlorogenic acid content \cite{coffee}. The worst explanation methods among those evaluated, GradientSHAP and a random explanation, shows either very small, non-contiguous  or randomly, scattered regions of interests and do not focus on parts of the time series that discriminate the two coffee types. 

\subsubsection{Video Motion Retrieval Dataset: GunPoint}
\begin{figure}[!htb]
\centering
  \includegraphics[width=1\textwidth]{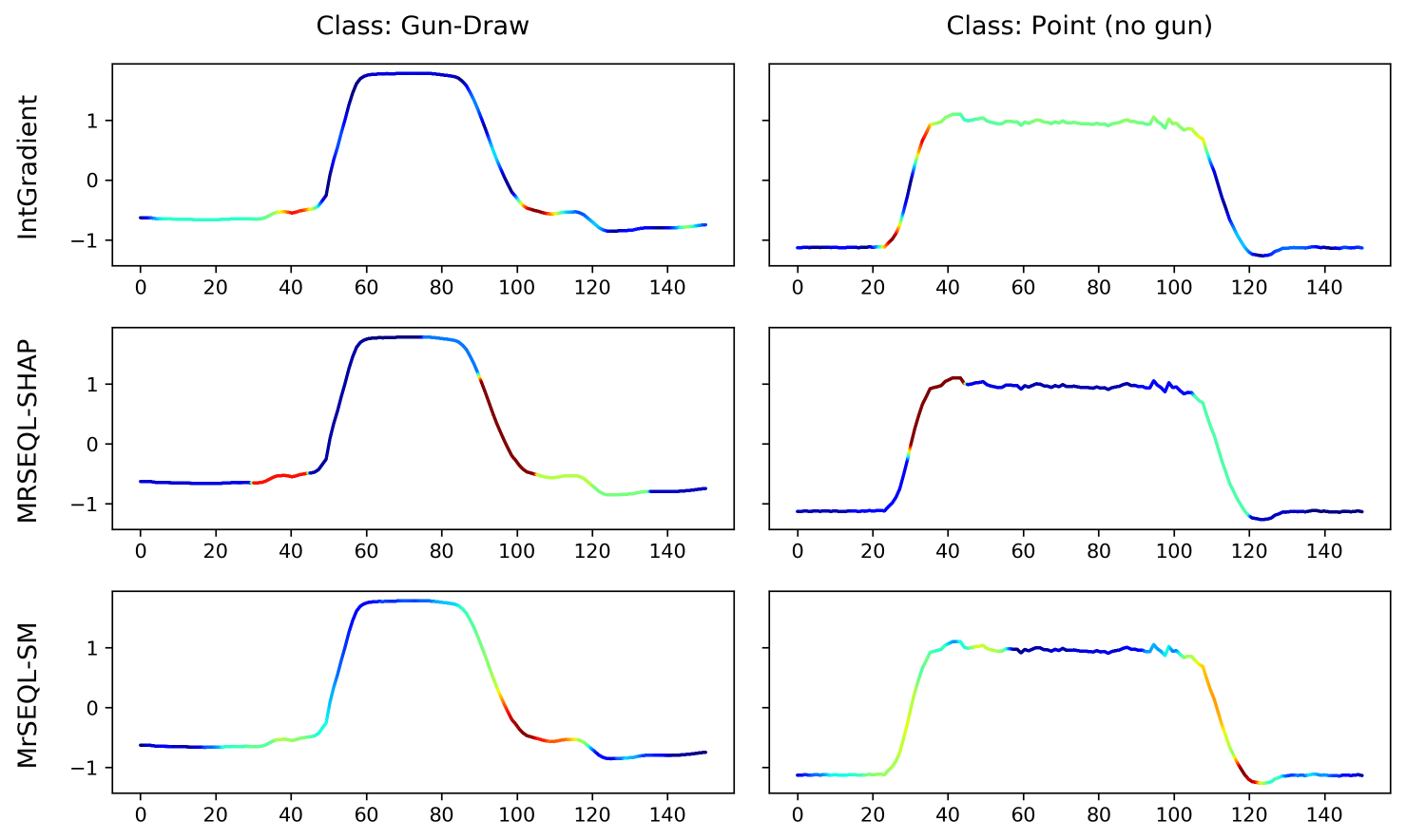}
\rule[1ex]{11.8cm}{1pt}
  
  \includegraphics[width=1\textwidth]{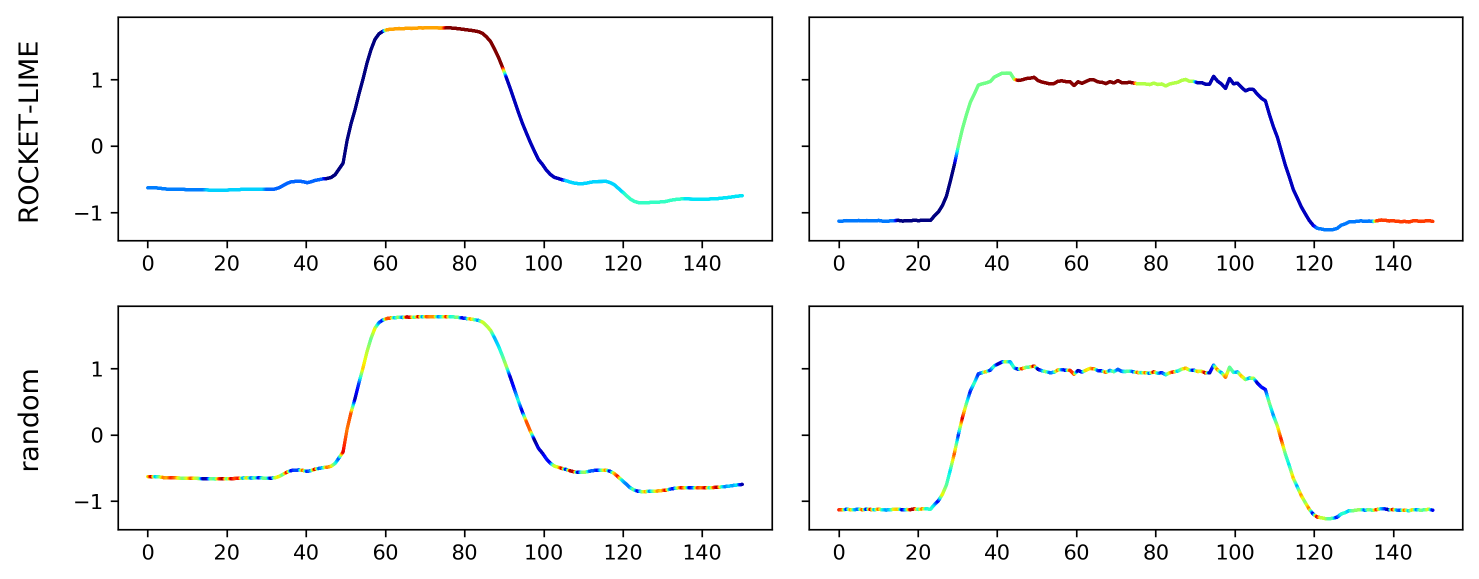}
  \caption{GunPoint Dataset: Visualization of top 3 best explanation methods (top 3 rows) and worst explanation method (bottom 2 rows) identified by AMEE. According to the description of the dataset in its original paper \cite{gunpoint},the discriminative region right before the high plateau of the two classes. For the Gun class, this region reflects the action of actors' hands moving above the holsters, and moving down to grasp the gun. For the Point class, there is no such action, resulting in a smoother curve from rest to point motion.}
  \label{fig:case-study-gunpoint}
\end{figure}

The famous GunPoint dataset is the time-series translation from a video sequence involving actors performing two distinct actions: pointing to a target with a gun (Gun class) and pointing with their index fingers only (Point class). This dataset was introduced in \cite{gunpoint} and is part of the UCR time series archive \cite{Dau2018TheArchive}.
Figure \ref{fig:case-study-gunpoint} visualizes the examples of explanations from the best three methods, worst method, and random method for this dataset. 
The expert ground truth for the GunPoint dataset conveys that the two classes differ in the steps where the Gun class requires the actor/actress to lift their hand above a holster, then reach down for the gun. This distinct action creates a subtle difference in the time steps right before the action of hands moving to the shoulder level (the sharp increase in time series values) to pointing the gun or hand (the plateau in the middle of the time series). The detailed description can be found in \cite{gunpoint}. 
In this dataset, AMEE identifies explanations from the IntegratedGradient method as the most computationally informative explanation, followed by MrSEQL-SHAP and MrSEQL-SM. The least informative method is ROCKET-LIME, which is even less informative than a random saliency explanation as this method refers to the wrong area of importance, failing to point out any of the salient regions of the time series.

\subsubsection{Motion Sensing Dataset: Counter Movement Jump (CMJ)}
\label{subsec:CMJ}

\begin{figure}[!htb]
\centering
  \includegraphics[width=1\textwidth]{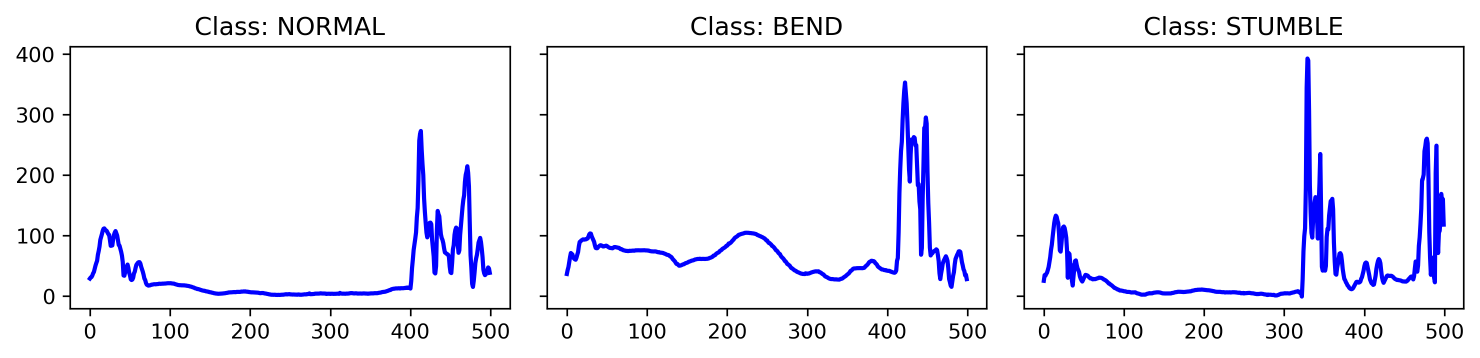}
  \caption{Examples of 3 classes of the Counter Movement Jump (CMJ) dataset.}
  \label{fig:case-study-vis}
\end{figure}

The CMJ dataset records the counter movement jumps of participants of 3 classes: Normal (jump done correctly), Bend (jump with knee bend), and Stumble (stumble at landing) (Figure \ref{fig:case-study-vis}).

\begin{figure}[!hbt]
\centering
  \includegraphics[width=1\textwidth]{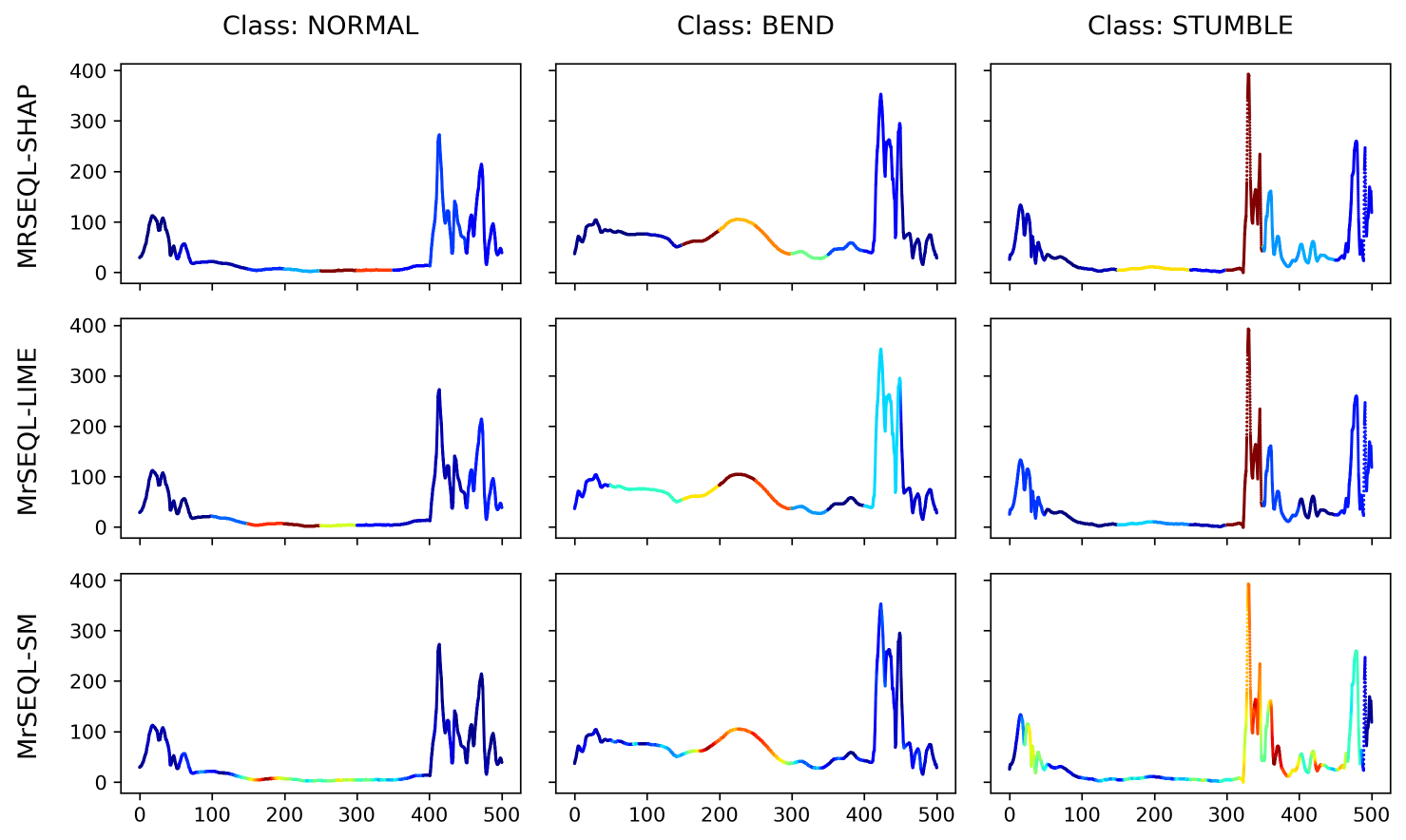}
  \rule[1ex]{11.8cm}{1pt}
  \includegraphics[width=1\textwidth]{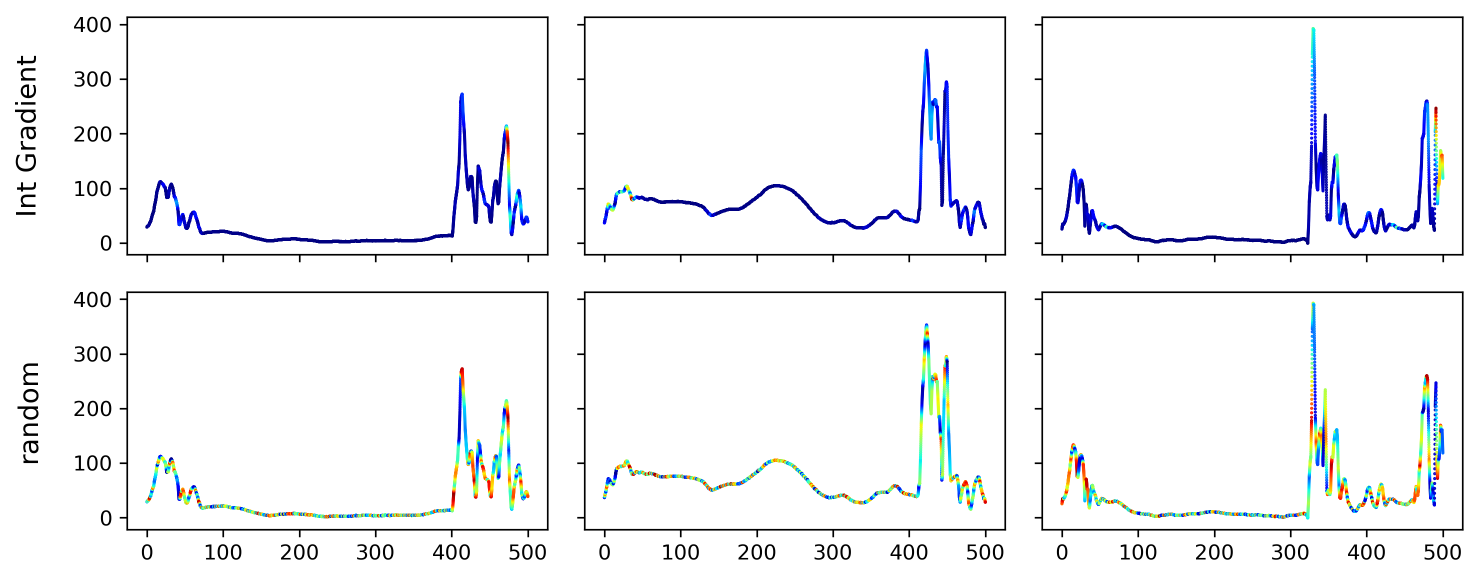}
  \caption{Counter Movement Jump (CMJ) dataset: Visualization of best explanation methods (top 3 rows), worst explanation method and random explanation (bottom 2 rows) identified by AMEE. Visualization of other explanations are presented in Section \ref{sec:appendix}.}
  \label{fig:case-study-selected}
\end{figure}

According to the domain experts who recorded this data \cite{LeNguyen2019InterpretableRepresentations}, the critical area for the first two classes  (NORMAL and BEND) is the middle part, while that of the final class (STUMBLE) is in the end of the time series. In class NORMAL, this region is completely flat.
The same region in class BEND is characterized by a hump in case participants' knees are in bending posture. In the STUMBLE class, the end of the time series is different from the previous two classes because of its very high, sharp peak due to a wrong landing position.

The result of AMEE for all studied  explanation methods is also given in Table \ref{tab:result_ucr}. The top 3 row show the top 3 explanations for this dataset are MrSEQL-SHAP (SHAP explanation based on MrSEQL classifier), MrSEQL-LIME (LIME explanation based on MrSEQL classifier), and MrSEQL-SM (saliency map obtained directly from MrSEQL classifier). We see a high agreement between these explainers as they all correctly highlight the corresponding discriminative areas provided by the expert (Figure \ref{fig:case-study-selected}). In addition, methods that are pointed out by AMEE as unreliable are also shown to highlight incorrect regions and do not agree with the opinion of the domain expert (e.g., explanation provided by Integrated Gradient method).

\subsubsection*{\textbf{Impact of Using Multiple Referees}}

\begin{figure}[p]
\begin{subfigure}{1\textwidth}
  \centering
  \includegraphics[width=1\linewidth]{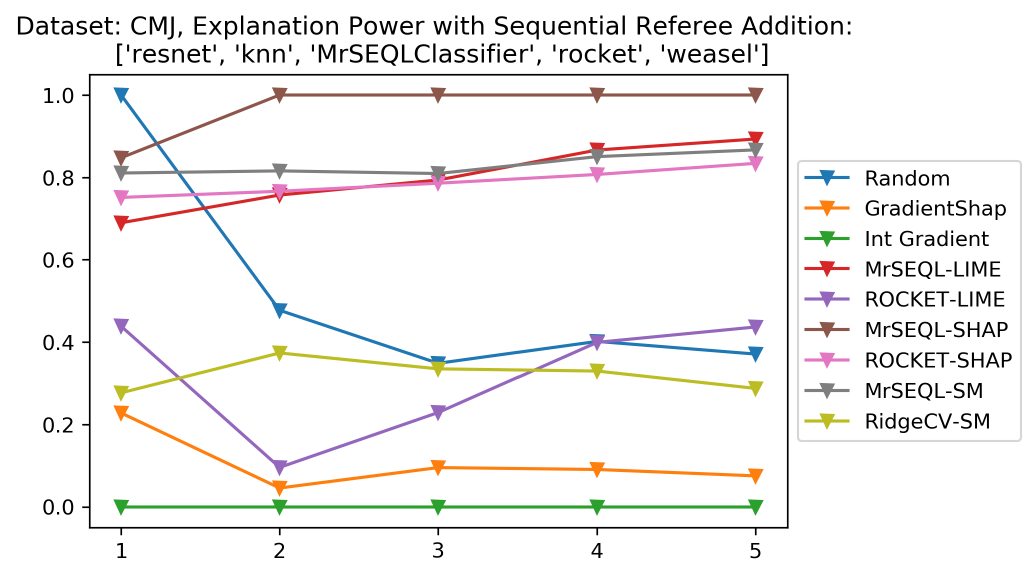}  
  \caption{Ensemble Referee - Sequential Addition}
  \rule[1ex]{11.8cm}{1pt}
\end{subfigure}
\begin{subfigure}{1\textwidth}
  \centering
  \includegraphics[width=1\linewidth]{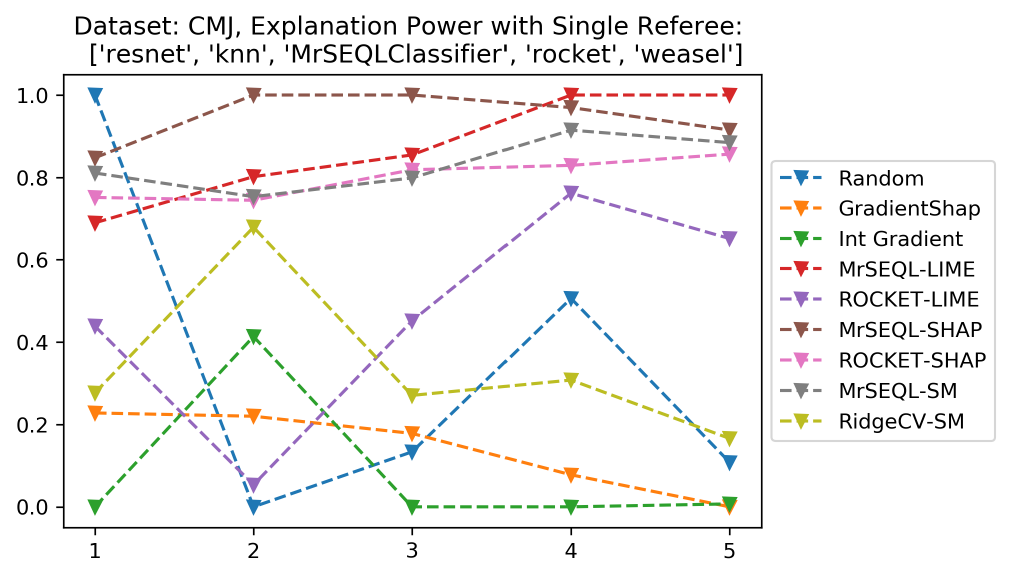}  
  \caption{Single Referee}
\end{subfigure}
\caption{CMJ dataset: Explanation Power in (a) Ensemble Referee Mode vs. (b) Single Referee Mode. The sequence of referees in both figures are (1) RESNET, (2) K-NN, (3) MrSEQL Classifier, (4) ROCKET, and (5) WEASEL 2.0.  In figure (a), Explanation Power is calculated in an ensemble approach with sequential addition of the referee classifiers. For example, the value of Explanation Power (reflected in the x-axis) of Random method (displayed in blue) when 2 classifiers are used (reflected in the y-axis) are aggregated from using both (1) RESNET and (2) K-NN. In figure (b), Explanation Power is calculated by using the evaluation from a single referee, without any aggregation from other referee. For example, the value of Explanation Power (reflected in the x-axis) of Random explanation (displayed in blue) when only the second classifier (in this figure, K-NN) is used (reflected in the y-axis). The dash is used to display the difference in Explanation Power in Figure (b) to reflect that there is no connection between the values of the explanation power between the evaluation results of the referee classifiers.}
\label{fig:sequential_vs_single_ref_CMJ}
\end{figure}

The Counter Movement Jump (CMJ) dataset is an example of a real dataset with known domain expert ground truth \cite{LeNguyen2019InterpretableRepresentations}. In our experiment, all of the referee classifiers achieve very high performance, ranging from 0.92 to 0.97 accuracy (Table \ref{tab:acc_ucr}.) 
Hence, this dataset presents an opportunity to investigate the benefit of using a committee of referee classifiers in comparison with using a single referee classifier. Figure \ref{fig:sequential_vs_single_ref_CMJ} shows the Explanation Power using two approaches: (a) using an ensemble of referees and (b) using a single referee independently. If we look at the case of Random explanation (displayed in blue) that should clearly be worse than the MrSEQL-based explanation (as shown in Section \ref{subsec:CMJ}). It is interesting that one of the referee classifiers that is quite accurate (Resnet with 0.92 accuracy) ranks the Random explanation as the best, with a significant Explanation Power difference to the others. If only a single referee was employed here, the recommendation could select the Random explanation. However, the risk decreases when an ensemble of referees are used. From this real example we observe that the benefit of using multiple referees is to improve the confidence and reliability of the evaluation, reducing the risk that a single referee is  wrong by instead aggregating evaluations from multiple referees.

%% file: 5discussion.tex
\subsection{Discussion}
\label{sec:discussion}

Our study carried over both synthetic and UCR datasets shows that AMEE can be used to computationally evaluate and rank different explanation methods.
We recommend the use of AMEE with full knowledge about the essential elements of the method. First, referees should be selected carefully, using classifiers of acceptable accuracy as determined by the application requirements. 
Using a committee of multiple accurate referee classifiers is recommended to reduce possible biases that one referee could introduce and results in a more reliable evaluation.
Second, having a variety of data perturbation methods is helpful, especially for hard-to-classify datasets. 
In addition, adding a random explanation while carrying out the evaluation with AMEE is helpful in identifying unreliable explanations. A worse-than-random explanation means that the explanation fails to trigger a change in referee classifiers when compared even to a random explanation, either not identifying the important areas, or not focusing on any important areas at all.
Finally, we recommend adding SHAP-based methods to accurate base classifiers for testing and further evaluation, as our experiments show that SHAP-based explanations often outperform other explanations using the same base classifiers. 

%% file: 5.1Recommendations.tex
\section{Recommendations for Practitioners}
\label{sec:recommendation}

In this section, we present our recommendations for using the AMEE framework to evaluate and recommend explanation methods. These are some of the lessons learned during the process of developing, designing, and conducting the experiments in this paper.
\begin{itemize}
    \item \textbf{Time Series Classifiers. } One of the key elements of our evaluation framework are  the referee classifiers. The more accurate the referees, the more reliable the result that we can potentially expect. Hence, choosing the right set of referees is a very important step before we even start to evaluate the explanation methods. We recommend using state-of-the-art time series classifiers that are well studied and compared in the latest empirical benchmarks  \cite{middlehurst2023bake}. When selecting  referees, we recommend to choose classifiers which are both accurate and computationally efficient, since  AMEE requires repeated inference of perturbed versions of the original dataset. 
    \item \textbf{Explanation Methods. } Unless the application users already have their preferred explanations pre-computed and only require AMEE for evaluation, we recommend to select explainers based on the extensive survey  \cite{Theissler2022}. We recommend to use diverse explanation methods, covering both intrinsic explanation and post-hoc explanation. From a computational perspective, we recommend to consider the cost of obtaining explanations. From our experience, explainers that use data segments (chunking) to explain time series seem  useful, but some of them are not efficient (for example, LIMESegment \cite{Sivill2022}). Additionally, we strongly recommend adding SHAP-based explanations to the list of explainers, as we observed these are highly informative in many datasets that we have tested. Finally, we recommend to add a random explanation to the evaluation (in addition to method-based explanations), for a simple sanity check. 
    \item  \textbf{Perturbation Strategy:} The perturbation strategy plays a critical role in both obtaining an explainer and for our recommendation framework. An effective perturbation strategy is one that, when used for perturbing the informative parts of the time series, leads to a change in prediction. In our experience, this effectiveness strongly depends on the specific dataset and classifiers, thus choosing the right perturbation can be tricky and time-consuming. Therefore, we recommend using multiple perturbation strategies in our framework.
    \item \textbf{Datasets \& Optimal number of Referees and Perturbation Strategies. } Our experiment covers a wide range of datasets of different classification difficulty level. We observe that when the dataset is easy to classify (e.g., many classification algorithms can achieve high accuracy), generally a lower number of referees and perturbation strategies can be used without affecting the evaluation results. However, when the dataset is harder to classify, using more referees and more perturbation methods is recommended to get a more reliable and less biased result. 
    \item \textbf{Adaptability.} We present AMEE as a robust explainer recommendation system for the Time Series Classification task. However, the framework is adaptable and could be generally applied to other types of data (such as images and text). For adapting to other data types, practitioners can consider more suitable perturbation methods and referee classifiers that work well with the target data. 
\end{itemize}

%% file: 6conclusion.tex
\section{Conclusion}
\label{sec:conclusion}

In this work we proposed AMEE, a Model-Agnostic Explanation Evaluation framework, for computationally assessing and ranking explanation methods for the time series classification task. 
We test the framework on 25 synthetic and  UCR archive datasets to obtain explanation evaluations for a wide variety of common explanation methods for time series, covering different aspects of explanation including type, scope and model dependency. 
Our experiments show a high agreement of the Explanation Power (measured by AMEE) in the synthetic datasets with the Oracle explanation (ground truth for each time point) and the Expert explanation in a real dataset (ground truth provided by a domain expert). We also find that  perturbation-based explainers based on SHAP generally perform better than gradient-based explainers for time series classification (given similar performance of the base models), but are computationally expensive.
AMEE can be used to select appropriate explanation methods for application users.  
It could also  potentially pinpoint inherent  problems, such as bias, that may exist in the training data and subsequently enhance the trustworthiness of AI systems in critical tasks.
This framework further empowers machine learning to discover new knowledge from the data. 
Another potential application is to use the best explainer recommended by AMEE as a proxy for downstream tasks to identify opportunities to compress data and optimize data storage, transmission, and analysis.
Finally, since AMEE relies on response to perturbation to evaluate importance of explanation methods, it can potentially be adapted to other types of data (such as images) and other machine learning tasks (such as time series regression and clustering). 
Future work includes devising a robust, AMEE-optimized, explanation method and using data experts to evaluate the validity and potential of knowledge discovery using this framework in biomedical and heathcare-related tasks such as genetic data understanding and sports analytics.

%% file: 0appendix.tex
\section{Appendix}
\label{sec:appendix}

\subsection{Explanation Power Calculation using Standard Scaler}

In Section \ref{subsec:metric}, we employ Min/Max Scaler for re-scaling the metrics in Step 2 and Step 4. While this standardization method is not the only available option, its advantages lies in the intuitive final metric in [0,1] range - with methods that are more informative would achieve a higher Explanation Power. Nevertheless, using other standardization method, such as Standard Scaling, is always an option to consider. Note that for this design choice, Step 5 is no longer logical and should be removed from the calculation (Algorithm \ref{algo:process_data_std_scaling}). The final metric can now be interpreted in a \textit{reverse} fashion to Explanation Power, with lower metric reflects a better explanation method. Table \ref{tab:result_synthetic_std} presents the results of the evaluation metric on the Synthetic datasets  using Standard Scaler as standardization method for Step 2 and 4, with an elimination of Step 5.

Results in Table \ref{tab:result_synthetic_std} shows a similar trend and agreement with Explanation Power presented \ref{tab:result_synthetic_std} and ground truth shown in Table \ref{tab:result_synthetic-f1-score}. For each dataset, methods with lowest values for the metric present in Table \ref{tab:result_synthetic_std} are associated with computationally most informative in explanation ability. For example, for SmallMiddle\_CAR dataset, random has highest metric value of 2.19 and associates with worst explanation method according to Table \ref{tab:result_synthetic-f1-score}. This result is similar with Table \ref{tab:result_synthetic_std}, in which this method has zero value in Explanation Power.

\begin{algorithm}
\small

\KwIn{
Set of XAI methods $M$, set of Perturbations $T$, set of Referees $R$, set of thresholds for important area $k$, test accuracy ($acc_{M,T,R,k}$)}
\KwOut{Average Scaled Explanation AUC ($asEAUC_M$), Average Scaled Rank ($asRank_M$), Explanation Power ($ePower_M$)}
Calculate Explanation AUC ($EAUC_{M,T,R}$) using $acc_{M,T,R,k}$ \\
Calculate rescaled AUC ($sEAUC_{M,T,R}$) by Standard Scaling $EAUC_{M,T,R}$ \\
Calculate Average Scaled Explanation AUC ($asEAUC_M$) of each $M$ by averaging $sEAUC_{M,T,R}$ across $R$ and $T$\\
Calculate Average Scaled Rank ($asRank_M$) of each $M$ by Standard Scaling $asEAUC_M$
\caption{Calculate comparison metric using Standard Scaling option}
\label{algo:process_data_std_scaling}
\end{algorithm}

 \setlength{\tabcolsep}{2pt}
 \renewcommand{\arraystretch}{1.1}
\begin{table}[!hbt]
    \centering
    \scalebox{0.75}{%
    \begin{tabular}{|l||c c c c c c c c c c|}
    \hline
   
        Dataset & Random & \shortstack{Grad\\SHAP} & 
        \shortstack{Int\\Gradient} & \shortstack{MRSEQL\\-LIME} & \shortstack{ROCKET\\-LIME} & \shortstack{MRSEQL\\-SHAP} & \shortstack{ROCKET\\-SHAP} & \shortstack{MRSEQL\\-SM} & \shortstack{RIDGECV\\-SM} & 
        \shortstack{Oracle} \\ \hline
       
        SM\_CAR             & 2.19       & -0.48 & -0.46 & 0.76         & 0.08         & -0.41       & -0.56       & 1.04   & -0.69   & -1.46 \\
SM\_NARMA           & 2.05       & -0.22 & -0.26 & 0.79         & 0.33         & -0.36       & -0.63       & 0.88   & -0.88   & -1.70 \\
SM\_Harmonic        & 2.01       & -0.82 & -0.81 & 1.15         & 0.24         & 0.66        & -0.48       & 0.28   & -0.95   & -1.27 \\
SM\_PseudoPeriodic  & 1.80       & -0.82 & -0.83 & 0.91         & 0.13         & 0.53        & -0.61       & 1.13   & -0.88   & -1.36 \\
SM\_GaussianProcess & 1.76       & 0.20  & -0.18 & 0.83         & -0.23        & 0.90        & -1.21       & -0.01  & -0.10   & -1.95 \\ \hline
        
        RT\_CAR             & 2.18       & -0.58 & -0.61 & -0.24        & -0.24        & -0.37       & -0.36       & 1.73   & -0.68   & -0.83 \\
RT\_NARMA           & 2.15       & -0.61 & -0.67 & -0.27        & 0.27         & -0.45       & -0.42       & 1.63   & -0.71   & -0.93 \\
RT\_Harmonic        & 2.15       & -0.84 & -0.87 & 1.11         & 0.04         & 0.36        & -0.56       & 0.55   & -1.07   & -0.88 \\
RT\_PseudoPeriodic  & 1.98       & -0.72 & -0.86 & 1.16         & -0.36        & 0.39        & -0.73       & 0.98   & -0.88   & -0.96 \\
RT\_GaussianProcess & 1.87       & -0.33 & -0.15 & 1.57         & -0.58        & 0.69        & -1.09       & -0.10  & -0.70   & -1.18 \\ 
    \hline        
    \end{tabular}
    }
    \smallskip
    \caption{Synthetic Datasets: Evaluation Metric using Standard Scaler for each of the 10 explanation methods evaluated on Synthetic datasets.}
    \label{tab:result_synthetic_std}
    
\end{table}

\subsection{Additional Tables and Figures}
\begin{figure}[htb]
\centering
  \includegraphics[width=1\textwidth]{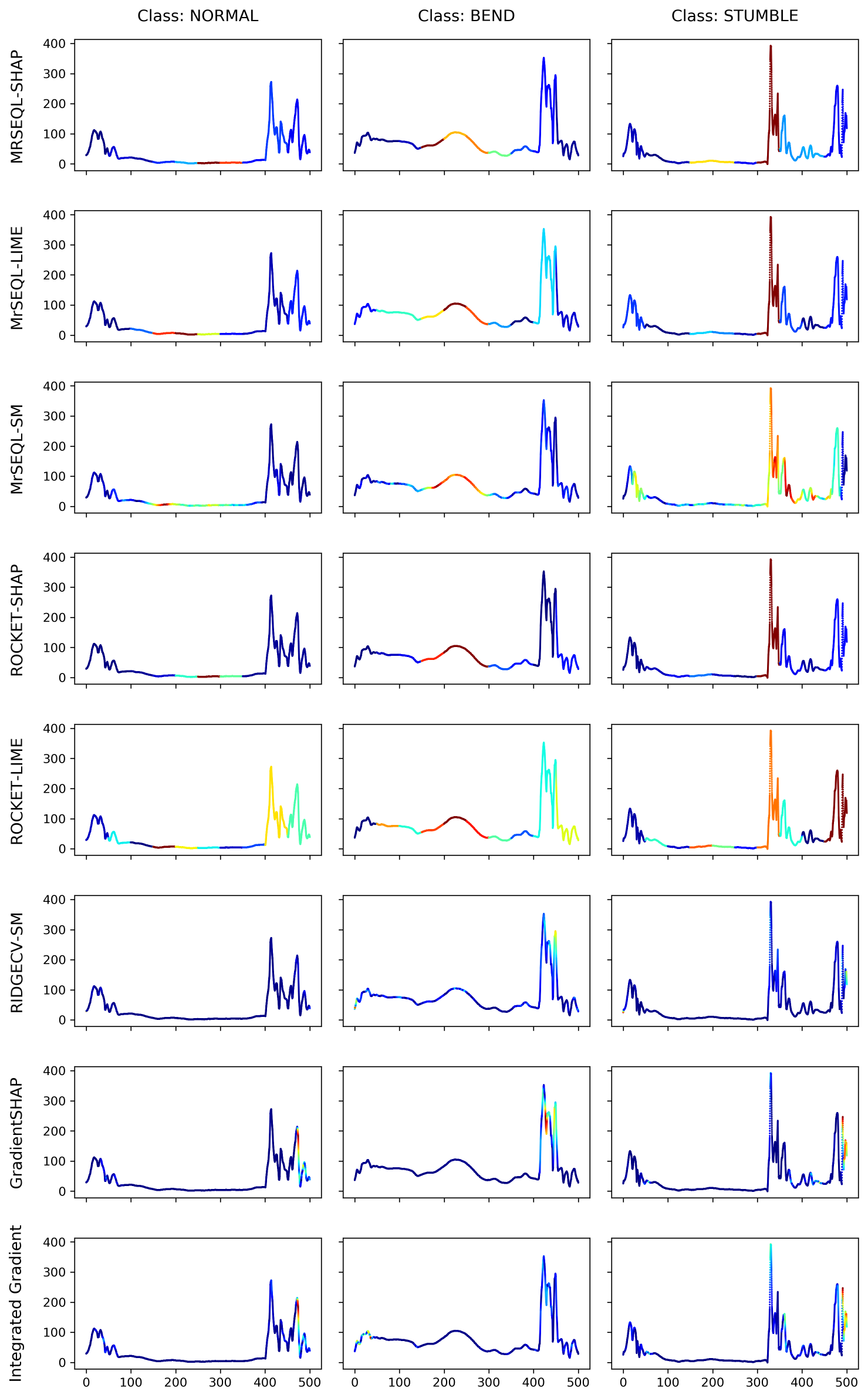}
  \caption{Visualization of all examined explanation methods on 3 classes of CMJ dataset (ordered by Explanation Power, high to low). This figure is best read in combination with results in Table \ref{tab:result_ucr}.}
  \label{fig:case-study-all}
\end{figure}

We include the full accuracy table for our experiments in Section \ref{sec:experiment} in Table \ref{tab:acc_synthetic} (Synthetic Data) and Table \ref{tab:acc_ucr} (Real Time Series Data). Figure \ref{fig:case-study-all} shows the visualization of all examined explanation methods on 3 classes of CMJ dataset in Section \ref{subsec:case-study}.

\begin{table}[!hbt]
    \centering
    \addtolength{\tabcolsep}{3pt} 
    \begin{tabular}{|l|ccccc|}
    \hline
        Dataset & MrSEQL & k-nn & RESNET & ROCKET & WEASEL \\ \hline
        SM\_CAR & \textbf{1.00} & \textbf{1.00} & \textbf{1.00} & \textbf{1.00} & \textbf{1.00} \\ 
        SM\_NARMA & \textbf{1.00} & \textbf{1.00} & \textbf{1.00} & \textbf{1.00} & \textbf{1.00} \\ 
        SM\_Harmonic & 0.85 & \textbf{1.00} & \textbf{1.00} & \textbf{1.00} & \textbf{1.00} \\ 
        SM\_PseudoPeriodic & 0.84 & \textbf{1.00} & \textbf{1.00} & \textbf{1.00} & \textbf{1.00} \\ 
        SM\_GaussianProcess & 0.64 & \textbf{0.94} & \textbf{0.84} & \textbf{0.89} & \textbf{0.85} \\ \hline
        RT\_CAR & \textbf{1.00} & \textbf{1.00} & \textbf{1.00} & \textbf{1.00} & \textbf{1.00} \\ 
        RT\_NARMA & \textbf{1.00} & \textbf{1.00} & \textbf{1.00} & \textbf{1.00} & \textbf{1.00} \\ 
        RT\_Harmonic & 0.80 & \textbf{1.00} & \textbf{0.96} & \textbf{1.00} & \textbf{0.97} \\ 
        RT\_PseudoPeriodic & 0.79 & \textbf{0.99} & \textbf{0.94} & \textbf{0.99} & \textbf{0.92} \\ 
        RT\_GaussianProcess & 0.51 & \textbf{0.87} & 0.64 & \textbf{0.80} & \textbf{0.73} \\ \hline
    \end{tabular}
    \smallskip
    \caption{Classifier accuracy on synthetic datasets. Classifiers that are selected as referees are in \textbf{bold}
    }
    \label{tab:acc_synthetic}
\end{table}

\begin{table}[!hbt]

    \centering
    \addtolength{\tabcolsep}{3pt} 
    \begin{tabular}{|l|ccccc|}
    \hline
        Dataset & MrSEQL & k-nn & RESNET & ROCKET & WEASEL \\ \hline
        ECG200 & \textbf{0.89} & \textbf{0.88} & 0.83 & \textbf{0.90} & \textbf{0.89} \\ 
        ECG5000 & \textbf{0.94} & \textbf{0.92} & \textbf{0.92} & \textbf{0.95} & \textbf{0.95} \\ 
        ECGFiveDays & \textbf{1.00} & 0.80 & \textbf{0.92} & \textbf{1.00} & \textbf{0.97} \\ 
        TwoLeadECG & \textbf{0.99} & 0.75 & \textbf{0.94} & \textbf{1.00} & \textbf{1.00} \\ \hline
        GunPoint & \textbf{0.99} & \textbf{0.91} & \textbf{0.97} & \textbf{1.00} & \textbf{1.00} \\ 
        CMJ & \textbf{0.96} & \textbf{0.92} & \textbf{0.92} & \textbf{0.97} & \textbf{0.97} \\ \hline
        PowerCons & 0.88 & \textbf{0.98} & \textbf{0.91} & \textbf{0.96} & \textbf{0.93} \\ \hline
        Coffee & \textbf{1.00} & \textbf{1.00} & \textbf{0.96} & \textbf{1.00} & \textbf{1.00} \\ 
        Strawberry & \textbf{0.96} & \textbf{0.95} & \textbf{0.94} & \textbf{0.98} & \textbf{0.98} \\ \hline
        Car & \textbf{0.85} & 0.73 & 0.40 & \textbf{0.90} & \textbf{0.92} \\ 
        ItalyPower & \textbf{0.91} & \textbf{0.96} & \textbf{0.95} & \textbf{0.97} & \textbf{0.96} \\ 
        Plane & \textbf{1.00} & \textbf{0.96} & 0.87 & \textbf{1.00} & \textbf{1.00} \\ 
        Sony1 & 0.75 & 0.70 & \textbf{0.89} & \textbf{0.92} & \textbf{0.94} \\ 
        Sony2 & 0.88 & 0.86 & 0.88 & \textbf{0.92} & \textbf{0.95} \\ 
        Trace & \textbf{1.00} & 0.76 & 0.75 & \textbf{1.00} & \textbf{1.00} \\   \hline
    \end{tabular}
    \smallskip
    \caption{Classifier accuracy on UCR datasets.     Classifiers that are selected as referees are in \textbf{bold}.
    \textit{Abbreviations:} CounterMovementJump - CMJ, ItalyPower - ItalyPowerDemand, Sony1/2 -SonyAIBORobotSurface1/2
    }
    
    \label{tab:acc_ucr}
\end{table}

